\documentclass[lettersize,journal]{IEEEtran}
\usepackage{graphicx}
\usepackage{amsmath,amsfonts}
\usepackage{algorithmic}
\usepackage{algorithm}
\usepackage{textcomp}
\usepackage{stfloats}
\usepackage{url}
\usepackage{verbatim}
\usepackage{cite}
\usepackage{subfig}
\usepackage{comment}
\usepackage{tabularx}
\usepackage{amsthm} 
\usepackage{enumitem} 
\usepackage{parskip}
\usepackage{amssymb}
\usepackage{booktabs}
\usepackage{multirow}
\usepackage{array}

\newtheorem{definition}{Definition}

\newtheorem{problem}{Problem}

\usepackage{bm}
\begin{document}

\title{Towards Effective Graph Rationalization via Boosting Environment Diversity}

\author{Yujie Wang, Kui Yu, Yuhong Zhang, Fuyuan Cao, Jiye Liang 
\thanks{Yujie Wang, Kui Yu, Yuhong Zhang are with the School of Computer Science and Information Engineering, Hefei University of Technology, Hefei, Anhui, China (email: yujiewang@mail.hfut.edu.cn; yukui@hfut.edu.cn; zhangyh@hfut.edu.cn).}
\thanks{Fuyuan Cao and Jiye Liang are with the School of Computer and Information Technology, Shanxi University, Taiyuan, China (email: cfy@sxu.edu.cn; ljy@sxu.edu.cn).}
\thanks{Manuscript received November 25, 2024; revised XX XX, XXXX.}}

\markboth{Journal of \LaTeX\ Class Files,~Vol.~XX, No.~X, August~XXXX}%
{Shell \MakeLowercase{\textit{et al.}}: A Sample Article Using IEEEtran.cls for IEEE Journals}


\maketitle
\begin{abstract}
Graph Neural Networks (GNNs) perform effectively when training and testing graphs are drawn from the same distribution, but struggle to generalize well in the face of distribution shifts. To address this issue, existing mainstreaming graph rationalization methods first identify rationale and environment subgraphs from input graphs, and then diversify training distributions by augmenting the environment subgraphs. However, these methods merely combine the learned rationale subgraphs with environment subgraphs in the representation space to produce augmentation samples, failing to produce sufficiently diverse distributions. Thus, in this paper, we propose to achieve an effective \underline{G}raph \underline{R}ationalization by \underline{B}oosting \underline{E}nvironmental diversity, a GRBE approach that generates the augmented samples in the original graph space to improve the diversity of environment subgraph. Firstly, to ensure the effectiveness of augmentation samples, we propose a precise rationale subgraph extraction strategy in GRBE to refine the rationale subgraph learning process in the original graph space. Secondly, to ensure the diversity of augmented samples, we propose an environment diversity augmentation strategy in GRBE that mixes the environment subgraphs of different graphs in the original graph space and then combines the new environment subgraphs with rationale subgraphs to generate augmented graphs. The average improvements of $7.65\%$ and $6.11\%$ in rationalization and classification performance on benchmark datasets demonstrate the superiority of GRBE over state-of-the-art approaches.

\end{abstract}

\section{Introduction}
\IEEEPARstart{G}{raph} Neural Networks (GNNs) are drawing widespread attention in various research fields, such as molecular property prediction, knowledge graph completion, etc \cite{liu2022pre}. However, existing GNNs are generally based on an independent and identically distributed (IID) assumption. They would show poor generalizability, i.e., suffer from a significant performance degradation on out-of-distribution (OOD) graphs \cite{li2022out}.

\begin{figure*}
	\centering
	\includegraphics[scale=0.75]{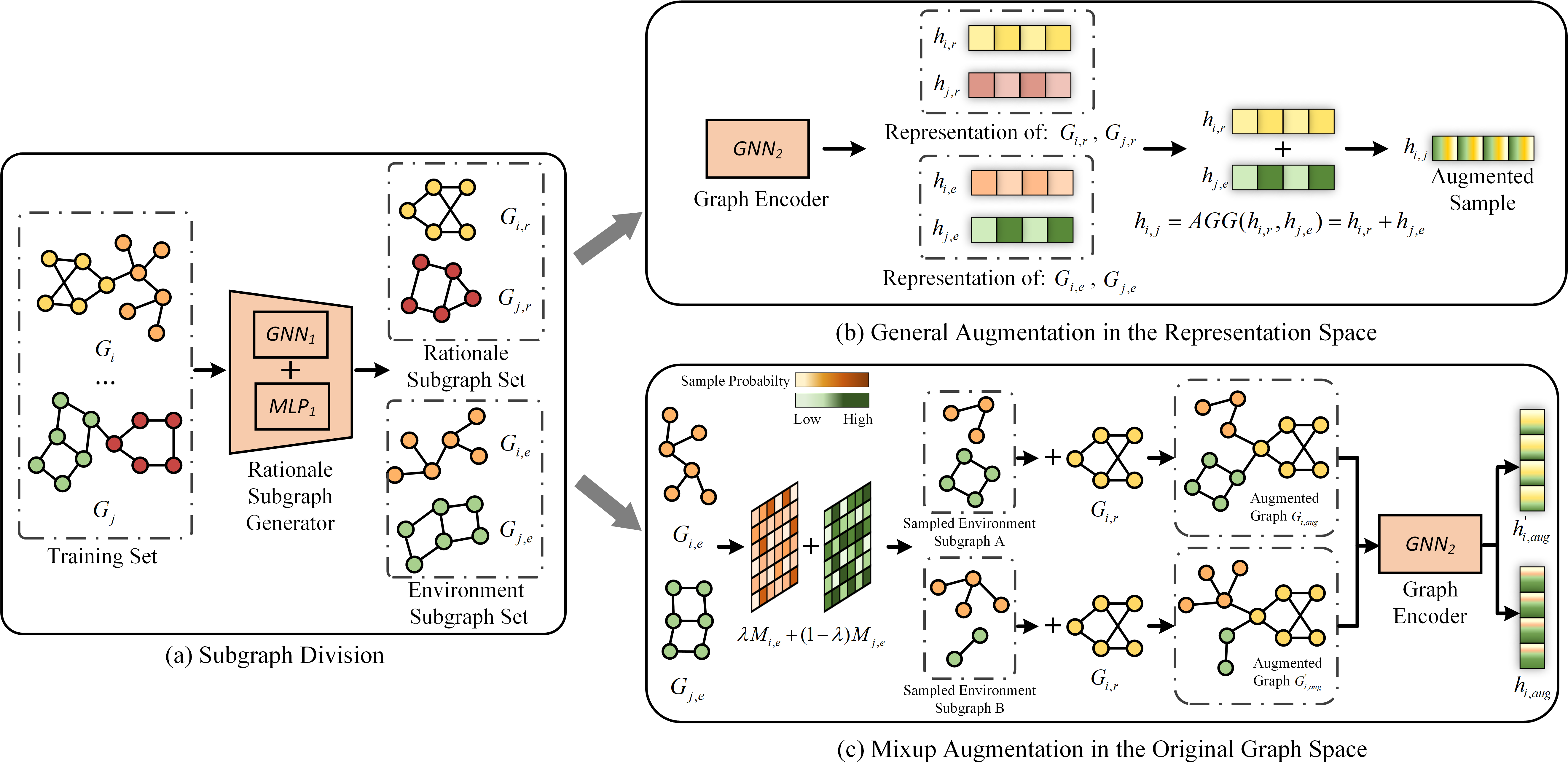}
	\caption{An example to illustrate the difference between the existing general augmentation in the representation space and our proposed mixup augmentation in the original graph space. Specifically, (a) Input graphs are divided into rationale and environment subgraphs. (b) General augmentation in the representation space splices the representations of rationale and environment subgraphs to create a new sample. (c) Our proposed Mixup augmentation operates in the original graph space by sampling environment subgraphs from a mixed distribution and attaching them to the rationale subgraph to generate diverse graphs.}
	\label{Example}
\end{figure*}

To address this OOD generalization issue, many research efforts \cite{miao2022interpretable, yang2023individual, jia2024graph, chen2022learning, gui2024joint, sui2022causal, li2022ood, yang2021learning, sun2024dive} have been proposed recently in the machine learning field. Among them, graph information bottleneck \cite{miao2022interpretable, yang2023individual} identifies a representative subgraph from the input graph that keeps the most predictive information of labels to make the model robust. Similarly, graph invariant learning \cite{jia2024graph, chen2022learning} applies a regularization term on multiple environments to capture the invariant relationships between subgraphs and labels. Unfortunately, most of these methods suffer from the lack of diverse training data to get a representative/invariant subgraph for stable prediction.
As another line of research, graph rationalization (GR) mostly enhances the diversity of training distribution by graph augmentation for generating new data samples \cite{wu2022discovering, liu2022graph, sui2022adversarial}.
As shown in Fig. \ref{Example} (a) and (b), the mainstream GR methods follow a two-step process, i.e., firstly identify rationale subgraphs and environment subgraphs, and then perform graph augmentation based on the learned rationale and environment subgraph representations to generate new samples for diversifying training data distribution. The rationale subgraph has deterministically predictive relationships with the labels and these relationships remain invariant across unknown distributions with shifts. In contrast, the environment subgraph is label-independent but may exhibit spurious correlations due to selection bias, which varies with changes in the distribution.

However, most current GR methods overlook the importance of subgraph segmentation accuracy, resulting in inaccurate subgraph divisions. Moreover, as shown in Fig. \ref{Example} (b), the diversity of augmented distributions produced by these methods remains constrained by the limited and unevenly distributed environment subgraph representations.

\begin{figure}[htbp]
    \centering
    \subfloat{\includegraphics[width=0.50\linewidth]{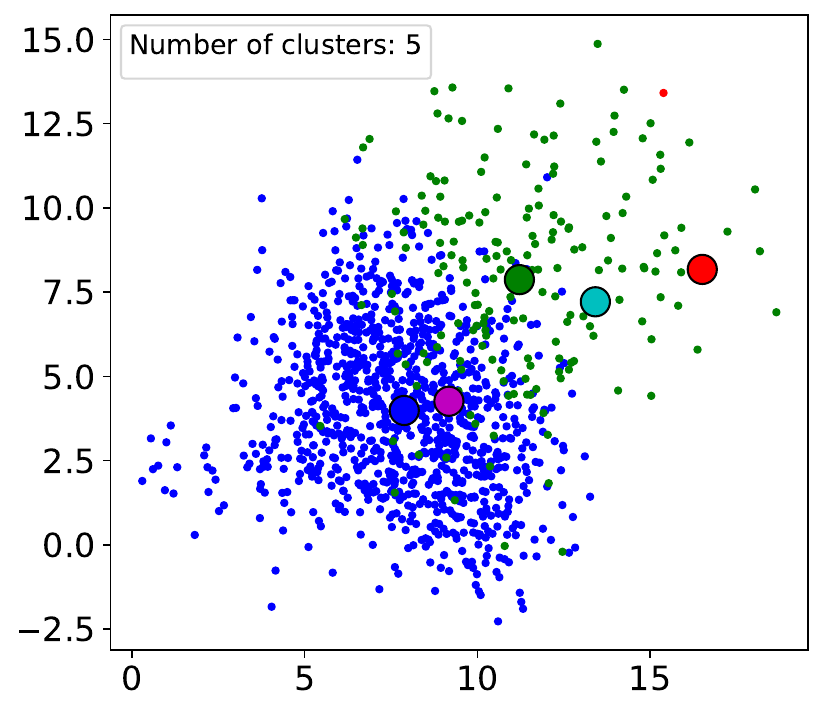}}
    \subfloat{\includegraphics[width=0.485\linewidth]{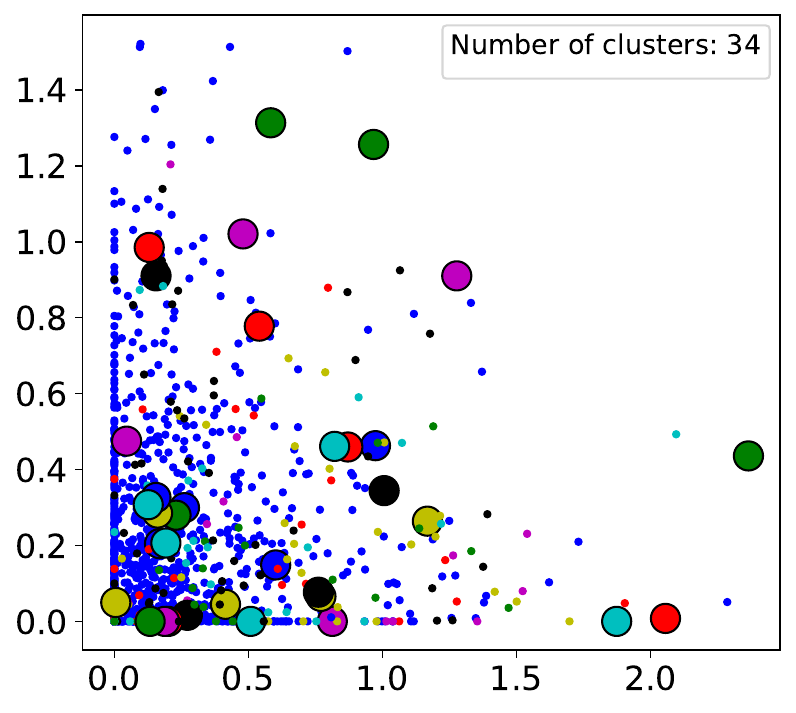}}
    \caption{The unsupervised clustering results of the environment subgraph representations on the Spmotif-0.9 dataset. (a) The representations are learned by the existing GREA method operated in the representation space. (b) The representations are obtained by our proposed GRBE method, which operates in the original graph space. Each node denotes an environment subgraph representation, and different colors represent different categories of environment subgraph representations.}
    \label{environment_cluster}
\end{figure}

For example, we select a state-of-the-art GR method called GREA \cite{liu2022graph} and utilize the mean shift clustering algorithm \cite{comaniciu2002mean} to visualize the representations of the environment subgraphs obtained by GREA on the Spmotif-0.9 \cite{ying2019gnnexplainer} and dataset. As illustrated in Fig \ref{environment_cluster}, GREA identifies five categories of environment subgraphs on the Spmotif-0.9 dataset, while this dataset has three categories of ground-truth environment subgraphs. Among these, the blue cluster accounts for the majority. Consequently, this leads to a higher likelihood of selecting the blue-corresponding environment subgraph representations when splicing the rationale and environment subgraph representations in the representation space to generate a new graph sample. Finally, the insufficient diversity in environment subgraph representation leads to a limited improvement in the diversity of the augmented distribution. This limitation arises because current augmentation methods simply combine rationale subgraph representations with existing environment subgraph representations in the representation space, without perturbing the environment subgraphs to generate more complex background information. Thus the insufficient diversity of environment subgraphs is still a bottleneck limiting the generalization ability of existing GR methods.

Therefore, instead of simply combining rationale subgraphs and environment subgraphs for augmentation in the representation space, in this paper, we explore graph augmentation in the original graph space by structure augmentation using edge removing and edge adding. Although this idea is promising to enhance the diversity of augmented samples, it faces the following two critical issues:

\textit{(1) How to improve the separation accuracy of rationale and environment subgraphs to ensure the effectiveness of augmentation?}
Ensuring the effectiveness of augmented graphs is the cornerstone of using them to find the optimal rationale. 
As pointed out by \cite{chen2024does}, the faithfulness of samples augmented by GREA remains questionable, as the augmentation is based on the inaccurate separation of the rationale and environment subgraphs.
Similarly, when performing augmentation in the original graph space, the labels of augmented graphs are determined by the rationale subgraph. However, this labeling process is sensitive to structure variation. Inaccurate subgraph partitioning may introduce false edges or nodes and even minor structure errors in the rationale subgraphs will damage the effectiveness of generated graphs. For instance, in the Spmotif dataset, the \includegraphics[height=1.1em]{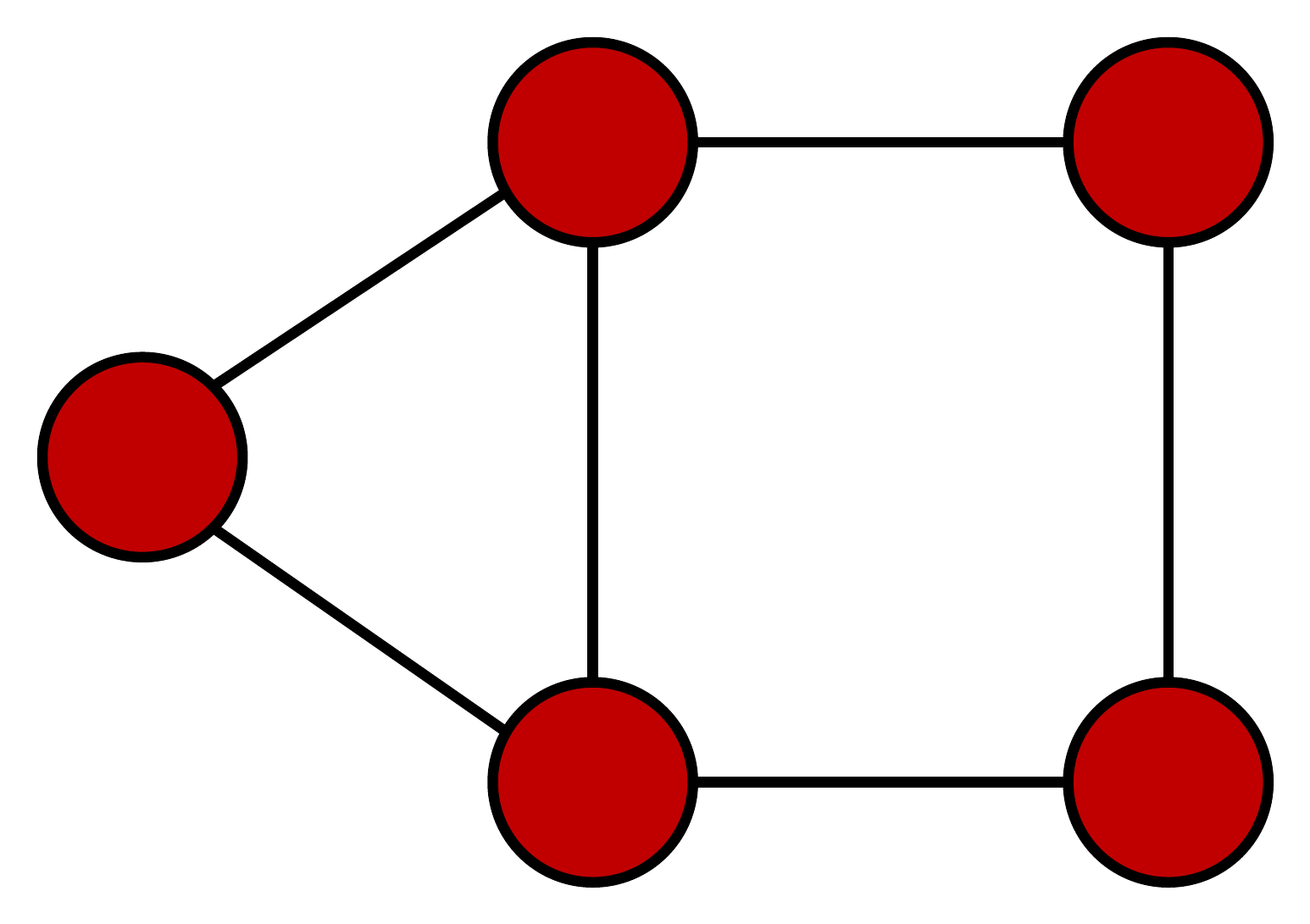} and \includegraphics[height=1.1em]{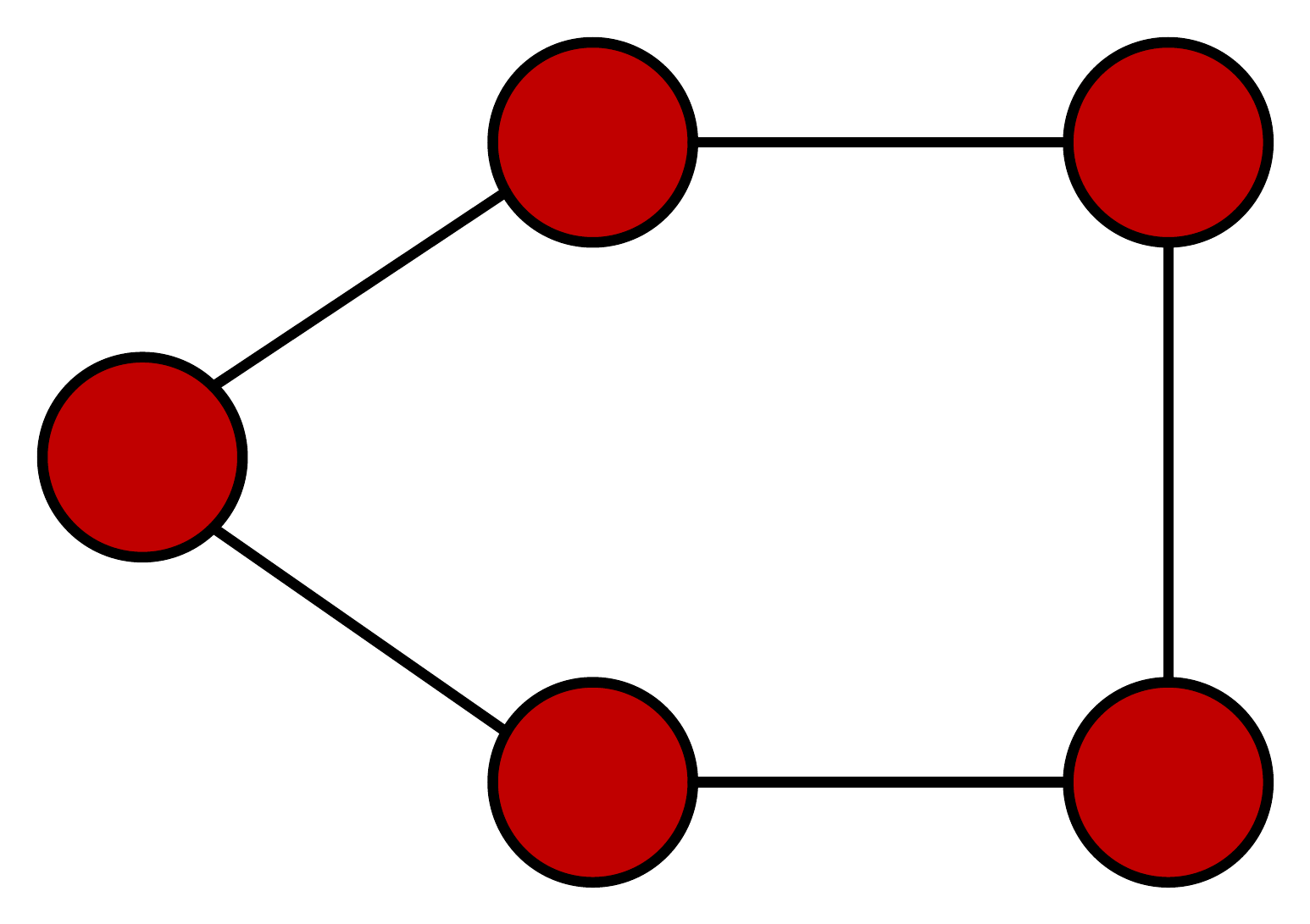} differ by merely one edge, but they represent two distinct labels: 'house' and 'cycle'. Thus, ensuring that rationale subgraphs capture invariant predictive relationships with the labels is crucial to subsequent augmentation.

\textit{(2) How to design an augmentation strategy that can generate reliable environment subgraphs to improve diversity?}
On the premise of ensuring effectiveness, another challenge is how to enhance the diversity and reliability of the augmented samples. In the original graph space, it is easy to generate sufficiently diverse environment subgraphs through structural modification. However, the augmented samples may be unreliable if the newly generated environment subgraphs are too randomized. That is to say, the newly generated environment subgraphs contain noisy topological structures that are irrelevant to the main tasks, which would make the distribution of the augmented samples deviate too much from the original distribution. Thus, it is essential to develop an effective augmentation strategy that can generate reliable samples to improve the distribution diversity of training graphs.

To address the above issues, we propose to augment graphs by boosting the diversity of the environment subgraph through a mixup augmentation strategy conducted in the original graph space. Fig. \ref{Example} shows the difference between our work and previous augmentation methods. Obviously, we can produce more diverse samples. This is because, compared with the augmentation in the representation space, our method is more fine-grained. From Fig. \ref{Example} (c), we can see that, by independently sampling nodes or edges in the existing environment subgraphs in the original space, we can decide whether each edge is sampled into the augmented environment subgraph and thus produce more diverse environment subgraphs. 
Then, as illustrated in Fig. \ref{environment_cluster}, our proposed mixup augmentation can produce 34 categories of environment subgraphs on the Spmotif-0.9 dataset, which is significantly more than the environment subgraphs learned by GRBE.
Overall, our work in this paper makes the following main contributions:

\begin{itemize}
	\item We achieve an effective graph rationalization by a boosting environment diversity method called GRBE, which consists of two mutually promoting modules: a Precise Rationale Subgraph Extraction (PRSE) module and an Environment Diversity Augmentation (EDA) module.
	\item To obtain a precise rationale subgraph, PRSE utilizes an adaptive rationale subgraph generation strategy and a self-supervised contrastive constraint to guide the GNN model in refining the rationale subgraph generation process.
	\item To generate reliable and sufficiently diverse augmented graphs, EDA presents a novel strategy that mixes the environment subgraphs of different graphs to generate new environment subgraphs and combines them with learned rationale subgraphs to generate new graph samples.
	\item Extensive experiments are conducted on multiple benchmark datasets with various distribution shifts, and the average improvement of $7.65\%$ and $6.11\%$ on rationalization and classification performance validate the effectiveness of our proposed method.
\end{itemize}

\section{Related Work}
Methods for improving the generalizability of the GNN model can be grouped into three main categories: graph information bottleneck, graph invariant learning, and graph rationalization.

\textbf{Graph information bottleneck} \cite{yu2021graph, yu2022improving, yang2024individual, miao2022interpretable, chen2022learning} adopts the information bottleneck principle to discover the representative subgraphs that determine graph labels. To name a few, GIB \cite{yu2021graph} designs a bi-level training scheme to identify maximally informative yet compact subgraphs. However, GIB tends to produce degenerated subgraph results due to the unstable bi-level optimization process. To address this issue, V-GIB \cite{yu2022improving} injects noise into the label-irrelevant parts of the original graph and approaches the compression condition of GIB, leading to a tractable variational upper bound. In IS-GIB \cite{yang2024individual}, the Individual Graph Information Bottleneck (I-GIB) applies explicit instance-wise constraints to eliminate spurious features in node- or graph-level representations, while the Structural Graph Information Bottleneck (S-GIB) emphasizes leveraging pairwise intra- and inter-domain correlations between node or graph instances to learn invariant features. Similarly, GSAT \cite{miao2022interpretable} injects stochasticity to the label-irrelevant graph components while learning masks with reduced stochasticity for the label-relevant subgraph. More recently, CIGA \cite{chen2022learning} proposes a causality model to capture the inherent invariance of graphs under various distributional shifts and adopts an information-theoretic objective to find the desired subgraphs for robust prediction. However, these methods merely focus on maximizing the predictive power of subgraphs for labels and do not eliminate spurious correlations, resulting in limited model generalization.

\textbf{Graph invariant learning} \cite{arjovsky2019invariant, ahuja2021invariance, krueger2021out, gui2024joint, yang2022learning, li2022learning, liu2023flood, jia2024graph, fan2022debiasing} aims to identify a subgraph that captures invariant relations with the labels across distinct environments.
For example, IRM \cite{arjovsky2019invariant} and its variant IB-IRM \cite{ahuja2021invariance}, VREX \cite{krueger2021out} adopt a regularization term to learn an invariant feature representation and further train a predictor that is simultaneously optimal in multiple environments. Gui et.al \cite{gui2024joint} proposes to adopt an adversarial training strategy to jointly optimize the label and environment causal independence for invariant subgraph discovery with theoretic guarantees. However, these methods rely on prior knowledge to divide the training data into various environments. In contrast, MoleOOD \cite{yang2022learning} does not require environment labels and realizes environment inference in an unsupervised manner. It proposes a learning objective that guides the graph encoder to utilize environment-irrelevant invariant subgraphs for stable prediction. MILI \cite{wang2024advancing} adopts a dual-head graph neural network with a shared subgraph generator to identify privileged substructures, and further leverages environment and task head to assign environment labels and learn invariant molecule representations. Similarly, GIL \cite{li2022learning} infers latent environments via clustering algorithms and devises a subgraph generator to generate invariant subgraphs. However, a limitation in GIL is that the graph encoder remains static after invariant learning, preventing flexible adaptation to target distributions. Thus, FLOOD \cite{liu2023flood} introduces a flexible bootstrapped learning component capable of refining the encoder during the testing phase.
However, the performance of most existing graph-invariant learning methods heavily relies on the diversity of training distributions.

\textbf{Graph rationalization} \cite{wu2022discovering, liu2022graph, yue2024learning, yu2023mind, yue2024cooperative} mostly enhances the diversity of training distributions by augmenting new data samples. For example, DIR \cite{wu2022discovering} firstly segregates the input graph into causal and non-causal subgraphs, and then it adopts causal intervention to generate multiple augmented distributions to find an optimal rationale subgraph. To mitigate the effects of the lack of sample problem on rationalization, GREA \cite{liu2022graph} proposes an environment replacement augmentation strategy, which generates new samples by replacing the environment subgraph representations and learns optimal rationale subgraphs on the real and augmented samples. More recently, SGR \cite{yue2024learning} proposes to learn precise rationale subgraphs by minimizing the mutual information between them and shortcut features, and then perform environment augmentation in the representation space to maintain a stable prediction. It is worth noting that the difference between our method and existing graph rationalization methods is that we perform environment subgraph augmentation in the original graph space. By adding/removing edges or nodes from the graph, our method can generate more diverse samples.
\section{Notations and Problem Definition}
We focus on finding graph rationales for improving the generalizability of graph classification tasks.
Let $\mathbb{G}$ and $\mathbb{Y}$ denote the graph and label space, respectively.
$G=(\mathcal{V}, \mathcal{E}; \bm{A}, \bm{X}) \in \mathbb{G}$ represents a graph, where $\mathcal{V}=\left\{v_1, v_2, \ldots, v_n\right\}$ is the set of nodes and $\mathcal{E} \in \mathcal{V} \times \mathcal{V}$ is the set of edges.
Here, $\bm{A} \in \mathbb{R}^{n\times n}$ is the adjacency matrix for $n$ nodes, where $A_{ij} = 1$ indicates an edge from node $v_i$ to node $v_j$, and $A_{ij} = 0$ otherwise.
The matrix $\bm{X} \in \mathbb{R}^{n \times d}$ contains the $d$-dimensional feature vectors for the nodes. Table \ref{notations} summarizes the important notations used in this paper.

\begin{table}[h]
    \centering
    \caption{The important notations used in this paper.}
    \resizebox{0.48\textwidth}{3.45cm}{
    \begin{tabular}{c|c}
        \toprule
        \midrule
        \textbf{Notation} & \textbf{Description} \\ 
        \midrule
        $\mathbb{G}$, $\mathbb{Y}$ & The graph and label space \\ \hline
        $G=(\mathcal{V}, \mathcal{E}; \bm{A}, \bm{X})$ & The graph with node set $\mathcal{V}$, edge set $\mathcal{E}$, \\
        $ $ & adjacency matrix $\bm{A}$, feature $\bm{X}$\\ \hline
        $G_r, G_e$ & The rationale and environment subgraph of $G$ \\ \hline
        $\bm{M}_r, \bm{M}_e$ & The sampling probability for $G_r$ and $G_e$\\ \hline
        $n$ & The number of nodes in $G$ \\ \hline
        $d$ & The dimension of feature vector $X$ \\ \hline
        $M_r^{uv}$ & The probability of edge $(u,v)$ being sampled into the $G_r$ \\ \hline
        $B_r^{uv}$ & The indicator represents whether edge $(u,v)$ is in the $G_r$ \\ \hline
        $G_{i,e}, G_{j,e}$ & The rationale subgraphs of the i-th and j-th graph\\ \hline
        $G_{i,e}, G_{j,e}$ & The environment subgraphs of the i-th and j-th graph\\ \hline
        $G^{p,+}$ & The positive sample pairs\\ \hline
        $G^{-}$ & The negative sample\\ \hline
        $t_p, t_n$ & The positive and negative augmentation function\\ \hline
        $\bm{A}_{ext}$ & The adjacency matrix of the environment subgraph \\
        $ $  & obtained by splicing $G_{i,e}$ and $G_{j,e}$ \\ \hline
        $\bm{M}_{mix,e}$ & The sampling probability of the mixed environment subgraph\\ \hline
        $M_{mix,e}^{uv}$ & The probability of edge $(u,v)$ being sampled from the $bm{A}_{ext}$ \\ \hline
        $B_r^{uv}$ & The indicator represents edge $(u,v)$ whether being sampled into the $G_r$ \\ \hline
        $G_{i,aug}$ & The new samples augmented by $G_r$ of the i-th graph \\ 
       \midrule
       \bottomrule
    \end{tabular}}
    \label{notations}
\end{table}

Following the existing work on rationalization \cite{wu2022discovering, liu2022graph, sui2022causal}, we consider each graph as consisting of two parts: the rationale $G_r = (\mathcal{V}_r, \mathcal{E}_r)$, which has stable predictive capability for the label $Y$, and the environment subgraph $G_e$, which is the complement of $G_r$ and has no causal relationship with $Y$. We formally define these subgraphs as follows:
\begin{definition}
Given an entire graph $G$ containing two subgraphs $G_r$ and $G_e$, if they satisfy the following conditions: (a) Complementarity: $G_r \cup G_e = G$ and $G_r \cap G_e = \emptyset$. (b) Sufficiency: $P(Y|G_r) = P(Y|G)$. (c) Independence: $Y \perp G_e | G_r$, then $G_r$ and $G_e$ are defined as the rationale and environment subgraph, respectively.
\label{definition1}
\end{definition}
Graph instances from $\mathbb{G}$ can be divided into a training set $D^{tr}=\{G^{e}_i,Y_i^{e}\}_{e\in \epsilon_{tr}}$ and a testing set $D^{te}=\{G^{e}_i,Y_i^{e}\}_{e\in \epsilon_{te}}$, where $\epsilon_{tr}$ and $\epsilon_{te}$ represent the training and testing environments, respectively. Since $\epsilon_{tr}$ and $\epsilon_{te}$ are sampled from different environments, a distribution shift exists between them due to the unknown selection bias in the data collection process, namely $P_{tr} \ne P_{te}$.
Based on the above notations and preliminaries, we formulate the problem of graph rationalization as follows.

\begin{problem}
A graph rationalization model $f:= \rho \: \circ \: h$ aims to find the rationale subgraph $G_r \subset G$ through a subgraph generator $h: \mathbb{G} \rightarrow \mathbb{G}_r$ and learn a classifier MLP $\rho: \mathbb{G}_r \rightarrow \mathbb{Y}$ to predict the graph label based on the $G_r$.
\label{problem1}
\end{problem}

\section{The Proposed GRBE Method}
\begin{figure*}
	\centering
	\hspace{5mm} 
	\includegraphics[scale=0.85]{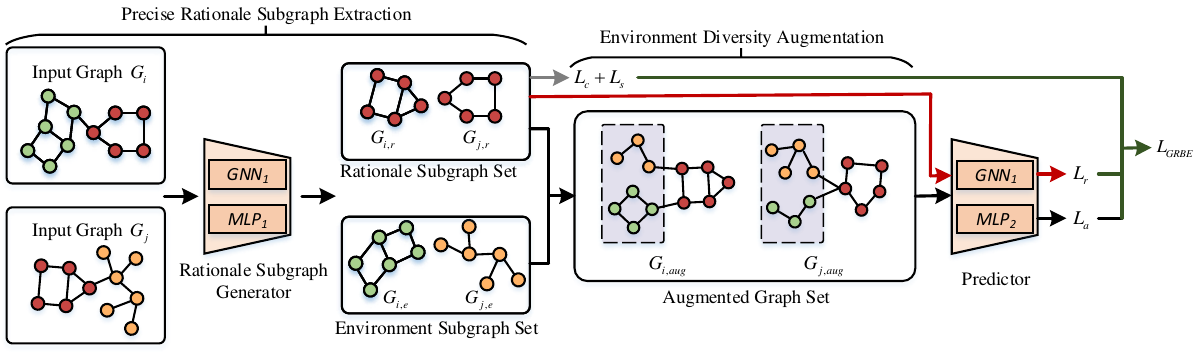}
	\caption{The overall framework of our proposed GRBE method. It performs environment diversity augmentation within the original graph space and can be divided into two main components: (a) PRSE guides the GNN model in learning a precise subgraph division. (b) EDA generates more diverse training graphs by exploring unknown environment subgraphs.}
	\label{framework}
\end{figure*}

Firstly, we present an overview of our proposed GRBE method. Fig. \ref{framework} shows the framework of GRBE, which includes two main modules: (1) PRSE adopts a contrastive constraint to guide the rationale subgraph generator achieving more precise subgraph division, which ensures the effectiveness of augmentation. (2) EDA mixes the probability distribution of environment subgraphs of different graphs and combines the sampled new environment subgraphs with the learned rationale subgraphs, which produces a more diverse training distribution.

\subsection{Precise Rationale Subgraph Extraction}
To obtain a more precise subgraph division, our PRSE strategy involves three stages: mask estimation, adaptive rationale subgraph generation and rationale subgraph refinement.

\textbf{Mask Estimation.} For an input graph $G$ with node set $\mathcal{V}$ and edge set $\mathcal{E}$, to separate the rationale and environment subgraphs from $G$, we employ a backbone $GNN_1$ to encoder the $G$ into node representations set $\{\bm{h}_v | v \in \mathcal{V}\}$. Subsequently, a Multilayer Perceptron $MLP_1$ is utilized to mask the concatenation $(\bm{h}_u, \bm{h}_v)$ of each edge $(u, v) \in \mathcal{E}$ into a scalar $M_{r}^{uv} \in [0, 1]$, which represents the probability that the current edge belongs to the rationale subgraph. The mask for the entire graph can be formulated as:
\begin{equation}
	\bm{M}_r=\sigma (MLP_1(GNN_1(G))).
	\label{mask}
\end{equation}
Here, $\sigma(\cdot)$ denotes sigmoid function. The items in complementary mask $\bm{M}_e = \bm{I}- \bm{M}_r$ indicate the probability of edges being classified into the environment subgraph, with $\bm{I}$ representing the identity matrix. It is worth noting that selecting rationale subgraphs can be facilitated by applying attention weights to nodes or edges. While our approach primarily focuses on edge weights, it can be readily extended to node-centric scenarios.

\textbf{Adaptive Rationale Subgraph Generation.} To generate the rationale subgraph $G_r$, existing methods \cite{wu2022discovering, chen2022learning} select the top-$K$ edges with the highest masks according to the $\bm{M}_r$.
However, implementing this approach presents two major issues: 1) Edges in the rationale subgraph may have a low value of mask and thus not be selected into the top-$K$ edges. 2) They impose a hard constraint on the size of the rationale subgraph. It is difficult to select a suitable value of $K$ to approximate the real proportion of the rationale subgraph without prior knowledge.

To address these two issues, we propose an adaptive rationale subgraph generation strategy without using $K$, which samples the edges of rationale subgraph from a Bernoulli distribution parameterized with $\bm{M}_r$. Firstly, in this way, edges with high mask weights have a high probability of being sampled into the rationale subgraph, while edges with low mask weights are also likely to be sampled.
Specifically, the adjacency matrix $\bm{A}_r$ of the rationale subgraph $G_r$ is identified as:
\begin{equation}
	\bm{A}_r = \bm{B}_r \odot \bm{A}, B_r^{uv} \sim \mathrm{Bern}(M_r^{uv}).
	\label{Bern_Sample}
\end{equation}
Here, $\odot$ is the element-wise product and $B_r^{uv}$ is the $(u,v)$-th entry of $\bm{B}_r$. If $B_r^{uv}=1$, current edge $(u,v)$ is sampled to the rationale subgraph, otherwise not. The remaining part of $G$ constitutes the environment subgraph $G_e$. Considering that the sampling process is non-differentiable, to make the gradient of $B_r^{uv}$ w.r.t $M_{r}^{uv}$ computable, we employ the concrete relaxation trick \cite{jang2017categorical} to estimate $B_r^{uv}$ by $\hat{B}_r^{uv}$:
\begin{equation}
	\hat{B}_r^{uv}=\operatorname{Sigmoid}\left(\frac{1}{t} \log \frac{M_r^{uv}}{1-M_r^{uv}}+\log \frac{u}{1-u}\right)
\end{equation}
where $t$ is the temperature parameter and $u \sim \text{Uniform}(0,1)$.

Secondly, we impose a soft constraint on the mask $\bm{M}_r$ to adjust the proportion of edges that are sampled into the rationale subgraph:
\begin{equation}
	\mathcal{L}_{s}=\left|\frac{1}{N_{\mathcal{E}}} \sum_{(u, v) \in \mathcal{E}} M_r^{uv}- r_s\right|
	\label{sparsity_loss}
\end{equation}
where $r_s \in [0,1]$ is the predefined sparsity level and $N_{\mathcal{E}}$ is the number of edges. Due to the inherent stochastic nature of the sampling process, the size of the sampled rationale subgraph can flexibly vary according to the $r_s$ and more closely approximate the ground-truth value compared to top-$K$ methods.

\textbf{Rationale Subgraph Refinement.}
In the previous step, initial rationale subgraphs are obtained. Due to the lack of supervision signals for the rationale subgraphs, existing rationalization models optimize the rationale subgraph learning process by minimizing the difference between the predicted labels of rationale subgraphs and the true graph labels. However, such an indirect optimization strategy often leads to inaccurate rationale subgraphs.
Unlike these models, we perform an additional refinement to the rationale subgraph learning process before predicting the labels, i.e., adopt self-supervised graph contrastive learning \cite{wei2023boosting} to directly guide the GNN towards better subgraph divisions. Graph contrastive learning captures invariant information by attracting positive sample pairs (i.e., augmented samples with the same labels) and repelling negative sample pairs (i.e., augmented samples with different labels). In the original space, it is easy to construct these pairs by selectively deleting nodes or edges in the graph. Conversely, constructing such pairs in the representation space is impossible because the rationale structure information that determines the labels is already vectorized via a message-passing mechanism.

Firstly, we construct positive sample pairs and maximize the similarity between these two positive samples, which can guide the adaptive rationale subgraph generation strategy described above capturing more edges that actually belong to the rationale subgraph. This is because, during the model training process, some edges in rationale subgraphs may be incorrectly assigned to environment subgraphs. This will result in a low similarity of the positive sample pairs constructed by perturbing the environment subgraphs while keeping the rationale subgraphs unchanged. By maximizing the similarity between the positive sample pairs, the model will give higher rationale subgraph sampling weights to the incorrectly assigned edges, thereby making the sampling results of the rationale subgraph more accurate.

To construct positive sample pairs $G^{p,+}$, we keep $G_r$ unchanged and perturb $G_e$ by applying node and edge dropout (any other graph augmentation method can be utilized here):
\begin{equation}
	G^{p,+} = t_p(G_e) + G_r, t_p \sim T_p
\end{equation}
where $T_{p}$ represents two different positive augmentation functions and $t_p$ denotes a sample from $T_p$. We then maximize the mutual information between the representation of these two positive samples:
\begin{equation}
	\label{Lp}
	\max_f \; I(\bm{h}^{1,+}, \bm{h}^{2,+}), \bm{h}^{p,+} = GNN_1(G^{p,+}).
\end{equation}
Here, $\bm{h}^{p,+}$ is the representation of the augmented positive samples and they are obtained through $GNN_1$.

However, solely relying on Eq. \ref{Lp} may cause the model to take more prominent environment information to easily achieve similarity in positive samples, leading to the learned rationale subgraph containing edges that actually belong to the environment subgraphs. To address this issue, we further generate negative samples $G^-$ to compress environment information in the learned rationale graph. By minimizing the similarity of negative sample pairs, the model is guided to reduce the weights of edges that are incorrectly assigned to the rationale subgraph. To construct negative samples $G^-$, for each $G$, we maintain its environment subgraph $G_e$ unchanged and perturb its rationale subgraph $G_r$ by performing a certain ratio of node and edge dropout:
\begin{equation}
	G^{-} = G_e + t_n(G_r)
\end{equation}
where $t_n$ is a negative augmentation function. Then, we minimize the mutual information between the negative sample pairs:
\begin{equation}
	\label{Ln}
	\min _f \; I(\bm{h}, \bm{h}^{-}), \bm{h} = GNN_1(G), \bm{h}^{-} = GNN_1(G^{-})
\end{equation}
where $\bm{h}$ and $\bm{h}^{-}$ are the representations of $G$ and $G^{-}$.
The overall objective of our contrastive learning is a Lagrangian of the Eq. \ref{Lp} and Eq. \ref{Ln}:
\begin{equation}
	\label{Lt}
	\min _f \; \mathcal{L}_{\mathrm{c}}: = -I(\bm{h}^{1,+}, \bm{h}^{2,+}) + I(\bm{h}, \bm{h}^{-}).
\end{equation}
Overall, the aim of this contrastive learning strategy is to enhance the model's ability to generate more precise rationale and environment subgraphs. When implementing the positive and negative augmentation function, we assign a sampling probability of 0.5 to each edge in the target subgraph, then traverse all the edges in the target subgraph and adopt the sampled subgraph as the result of augmentation.

\subsection{Environment Diversity Augmentation}
In Section 4.1, we obtained a refined rationale and environment subgraph division and ensured the effectiveness of augmented samples. Next, we introduce our proposed EDA strategy that generates more reliable and diverse training graphs in the original graph space. EDA involves two key steps: environment diversity exploration and augmented graph synthesis.

\textbf{Environment Diversity Exploration.}
The primary goal of environment diversity exploration is to generate environment subgraphs absent from the training dataset. These generated environment subgraphs can reduce the distribution shifts between the training set and testing set, and thus help GNN models generalize effectively under distribution shifts.
To achieve this goal, existing graph data augmentation methods \cite{ suresh2021adversarial, zhao2021data, park2022graph} modify the topology structures by randomly adding or removing edges.
However, such a random modification method would introduce task-irrelevant structures and make the augmented graphs unreliable \cite{wei2023boosting}, causing the distribution of augmented graphs to deviate excessively from the original distribution.

Thus, it is natural to consider concatenating two randomly selected environment subgraphs in the original space to form new environment subgraphs. However, such an operation would introduce two major issues:
1) Random selection has a high probability of getting a dominant environment subgraph. 2) The two selected environment subgraphs may be in the same category. Both of these issues will lead to a fatal consequence, that is, the improvement of distribution diversity by augmented samples is limited because the categories of augmented environment subgraphs remain the same as those of the existing environment subgraphs.
To address these two issues, we propose an environment subgraph augmentation strategy based on mixup, which mixes the subparts of existing environment subgraphs to produce a new category of environment subgraph.
For example, \includegraphics[height=1.25em]{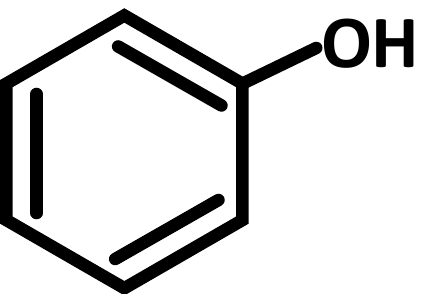} and  \includegraphics[height=1.25em]{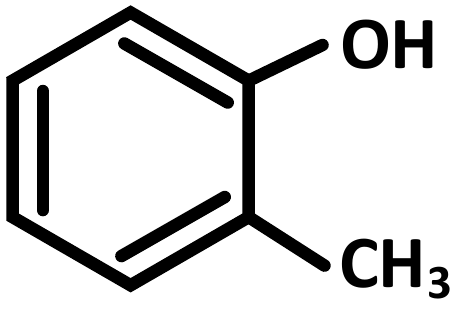} represent two phenolic compounds. If we mix the two benzene rings in them, we get a new biphenyl, \includegraphics[height=1.1em]{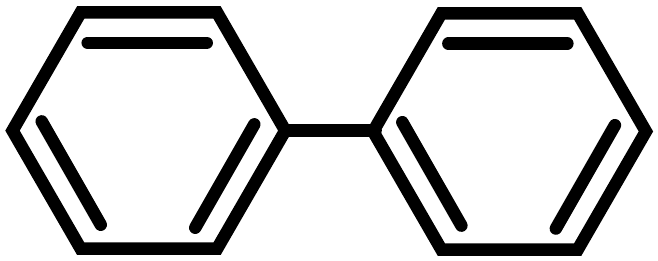}, which is not a phenolic compound. Specifically, the details of our proposed environment subgraph augmentation strategy are described as follows.

Firstly, we randomly select two graphs $G_i, G_j$, and obtain their corresponding environment subgraphs $G_{i,e}, G_{j,e}$ according to the Eq.\ref{Bern_Sample}. We represent the adjacency matrix of $G_{i,e}, G_{j,e}$ as $\bm{A}_{i,e}, \bm{A}_{j,e}$. To mix two graphs with different sets of nodes, we first extend node sets in $G_{i,e}$ and $G_{j,e}$ into a single node set and then mix their corresponding masks $\bm{M}_{i,e}$ and $\bm{M}_{j,e}$ to generate the mask of new environment subgraph:
\setlength{\arraycolsep}{3pt}
\begin{equation}
	\bm{A}_{e x t}=\left[\begin{array}{cc}
		\bm{A}_{i, e} \, & \mathbf{0} \\
		\mathbf{0} \, & \bm{A}_{j, e}
	\end{array}\right], \bm{M}_{m i x, e}=\left[\begin{array}{cc}
		\lambda \bm{M}_{i, e} & \mathbf{0} \\
		\mathbf{0} & (1-\lambda) \bm{M}_{j, e}
	\end{array}\right].
	\label{mix}
\end{equation}
Here, $\bm{A}_{ext}$ is the adjacency matrix of the extended graph. $\bm{M}_{mix,e}$ is the sampling probability distribution of the new environment subgraph, the $(u, v)$-th entry of the $\lambda \bm{M}_{i,e}$ represents the edge $(u, v)$ in $\bm{A}_{i,e}$ has a $\lambda \bm{M}_{i,e}^{u,v}$ probability being sampled into the new environment subgraph, which is similar in the $(1-\lambda) \bm{M}_{j, e}$. 

Secondly, based on the mask obtained in Eq. \ref{mix}, we sample a portion of edges from each of the two existing environment subgraphs $G_{i,e}, G_{j,e}$ to form a new environment subgraph $G_{mix,e}$, whose category may be different from the previous environment subgraphs. Specifically, similar to the sampling process in Eq. \ref{Bern_Sample}, the adjacency matrix of $G_{mix,e}$ is:
\begin{equation}
	\bm{A}_{m i x, e}=\bm{B}_{m i x, e} \odot \bm{A}_{e x t}, B_{m i x, e}^{u v} \sim \operatorname{Bern}\left(M_{m i x, e}^{u v}\right)
\end{equation}
where $M_{mix,e}^{uv}$ is the $(u, v)$-th entry of the $\bm{M}_{m i x, e}$ defined in Eq. \ref{mix},  $B_{mix,e}^{uv}\in[0,1]$ determines whether the current edge $(u, v)$ is sampled to the newly generated environment subgraphs $G_{mix,e}$. We adopt the concrete relaxation trick again to make the gradient w.r.t $M_{mix,e}^{uv}$ computable. In the augmentation process, $\lambda \in [0,1]$ is a hyperparameter that controls the contribution from the environment subgraphs $G_{i,e}$ and $G_{j,e}$. If $\lambda$ approaches 0 or 1, the sampled new environment subgraph will be similar to $G_{i,e}$ or $G_{j,e}$, and the improvement of distribution diversity is limited. By adjusting $\lambda$, we can flexibly control the diversity of the augmented distribution.

\textbf{Augmented Graph Synthesis.} Here, we have obtained the rationale and diverse environment subgraphs. Now, we have to synthesize augmented graphs. Specifically, for a graph $G_i$, we preserve its rationale subgraph $G_{i,r}$ and replace its own environment subgraph $G_{i,e}$ with the generated environment subgraph $G_{mix,e}$.  According to the definition \ref{definition1}, the label of a graph is determined by its rationale subgraph. Thus, the label of $G_{mix,e}$ remains the same as the label of $G_i$. Meanwhile, to ensure connectivity between the rationale and environment subgraphs so that message passing mechanism can perform well, we add $N_{add}$ edges between these two subgraphs according to the node degree, where $N_{add} = r_{add} *(|E_{G_i}| + |E_{G_j}|)$ and $r_{add}$ is the pre-defined ratio (e.g. 10\%) and $|E_{G_i}|, |E_{G_j}|$ are the numbers of edges in the $G_i, G_j$ respectively. Finally, we obtain the augmented samples:
\begin{equation}
	G_{i, aug} = G_{i,r} + G_{mix,e}, i=1,2,3...N_{aug}
\end{equation}
where $N_{aug} = r_{aug} * |G|$ is the number of augmented samples and $|G|$ is the number of training graphs. Inspired by \cite{wang2024do}, we use augmentation ration $r_{aug} \in [0,1]$ to balance the complexity and performance of augmentation.

\subsection{Optimization}
Based on the augmented graphs obtained after EDA,  we utilize both the original and augmented graphs to optimize the graph rationalization model in an end-to-end way.
Firstly, as the rationale subgraph is regarded as the invariant property of the graph label, it should be good for label prediction. Thus, we predict the label of $G_i$ based on the extracted rationale subgraph $G_{i,r}$ :
\begin{equation}
	\hat{y}_{i,r} = MLP_{2}(\bm{h}_{i,r})
	\label{rationale_prediction}
\end{equation}
Meanwhile, to constrain the distribution difference between the newly augmented samples and the original samples, we also predict the label of the augmented graph $G_{i,aug}$:
\begin{equation}
	\hat{y}_{i,aug} = MLP_{2}(\bm{h}_{i,aug})
	\label{augmentation_prediction}
\end{equation}
Here, the graph representation of rationale subgraphs and augmented graphs, denoted as $\bm{h}_{i,r}$ and $\bm{h}_{i,aug}$ are encoded by the $GNN_1$ used in the subgraph generation process. They are then fed into the same predictor $MLP_2$ to predict the label.
We use the CrossEntropy function to compute the predictive loss of rationale subgraphs and augmentation graphs as $\mathcal{L}_r$ and $\mathcal{L}_a$. Without loss of generality, we focus on the binary classification tasks, the predictive loss of a batch with $N$ graphs is defined as:
\begin{equation}
	\mathcal{L}_{r}=\frac{1}{N} \sum_{i=1}^N\left(y_i \cdot \log \hat{y}_{i}+\left(1-y_i\right) \cdot \log \left(1-\hat{y}_{i}\right)\right), 
\end{equation}
which is similar for the $\mathcal{L}_a$ and the $\hat{y}_{i}$ can be be instantiated by Eq. \ref{rationale_prediction} and Eq. \ref{augmentation_prediction}.

For the mutual information in Eq. \ref{Lt}, we adopt InfoNCE \cite{chen2020simple} as the estimator. Given a training batch with $N$ graphs, for arbitrary $G^{\prime}, G^{\prime \prime}$ and its corresponding representation $\bm{h}^{\prime}=GNN_1(G^{\prime}), \bm{h}^{\prime \prime}=GNN_1(G^{\prime \prime})$, $I(\bm{h}^{\prime}, \bm{h}^{\prime \prime})$ can be estimated as:

\begin{equation}
    \footnotesize
	\hat{I}(\bm{h}^{\prime}, \bm{h}^{\prime \prime})= \frac{1}{N} \sum_{i=1}^N \log \frac{\exp \left({\bm{h}^{\prime}}_i^T{\bm{h}^{\prime \prime}}_i / \tau\right)}{\exp \left({\bm{h}^{\prime}}_i^T{\bm{h}^{\prime \prime}}_i / \tau\right)+\sum_{j=1, j \neq i}^N \exp \left({\bm{h}^{\prime}}_i^T{\bm{h}^{\prime \prime}}_j / \tau\right)} .
\end{equation}

where $G^{\prime}, G^{\prime \prime}$ are the positive and negative sample pairs in Eq. \ref{Lt}, and $\tau$ denotes the temperature parameter.

Finally, combining the above loss and the sparsity loss defined in Eq. \ref{sparsity_loss}, the total objective function of our GRBE is defined as follows:
\begin{equation}
	\arg \min_{f} \mathcal{L}_{GRBE} = \mathcal{L}_{\mathrm{r}} + \alpha \mathcal{L}_{\mathrm{a}} + \beta \mathcal{L}_{\mathrm{c}} + \gamma \mathcal{L}_{\mathrm{s}}
\end{equation}
where $\alpha, \beta, \gamma$ are three hyperparameters that balance the contribution of the different items. In the inference process, only $G_r$ is generated to yield the predictive results.
\section{Experiments}

\subsection{Experiment setting}
\subsubsection{\textbf{Datasets}}
We conduct experiments on the synthetic dataset: Spurious-Motifs \cite{ying2019gnnexplainer} and real-world datasets: Open Graph Benchmark (OGBG) \cite{hu2020open}, Graph-SST2/SST5 \cite{socher2013recursive}, Twitter\cite{socher2013recursive}, MUTAG \cite{rupp2012fast}. We give more details about the synthetic and real-world datasets and show the statistics of the above datasets in Table \ref{dataset}.

\begin{table}[htbp]
	\centering
	\caption{The detailed statistic of datasets.}
	\resizebox{0.48\textwidth}{2.275cm}{
		\begin{tabular}{c|ccccc}
			\toprule
			\midrule
			Dataset & graphs & avg.nodes & avg.edges & classes & features \\
			\midrule
			Spmotif-0.9 & 18000 & 29.6  & 42    & 3   & 4 \\
			Spmotif-0.7 & 18000 & 30.8  & 45.9  & 3   & 4 \\
			Spmotif-0.5 & 18000 & 29.4  & 42.5  & 3   & 4 \\
			\midrule
			OGBG-Molbace     & 1513  & 34.1  & 73.7  & 2   & 9 \\
			OGBG-Molbbbp     & 2021  & 23.9  & 51.9  & 2   & 9 \\
			\midrule
			Graph-SST2  & 43779 & 13.7  & 25.3  & 2   & 768 \\
			Graph-SST5  & 9516  & 19.85 & 37.7  & 5   & 768 \\
			Twitter     & 5441  & 21.1  & 40.2  & 3   & 768 \\
			\midrule
			MUTAG       & 188   & 17.8  & 19.6  & 2    & 14 \\
                \midrule
			\bottomrule
	\end{tabular}}
	\label{dataset}
\end{table}

\begin{table}[htbp]
	\centering
	\caption{The Detailed parameterization of GRBE on different datasets.}
	\resizebox{0.48\textwidth}{2.5cm}{
		\begin{tabular}{c|ccc|ccccc}
			\toprule
			\midrule
			& \# hiddens & \# layers & epochs & $\alpha$ & $\beta$ & $\gamma$ & $r$  & $r_s$ \\
			\midrule
			Spmotif-0.9   & 256  & 3 & 200  & 0.5  & 0.1  & 0.5  & 0.2 & 0.7 \\
                \midrule
			Spmotif-0.7   & 64  & 5 & 200 & 0.5  & 0.1  & 0.5  & 0.5 & 0.7 \\
                \midrule
			Spmotif-0.5   & 64  & 5 & 200 & 0.5  & 0.1  & 0.5  & 0.2 & 0.7 \\
                \midrule
			Graph-SST2   & 256  & 3 & 30 & 0.5  & 0.1  & 0.5  & 0.2 & 0.7 \\
                \midrule
			Graph-SST5   & 128  & 3 & 100 & 1  & 0.1  & 0.1  & 0.2 & 0.7 \\
                \midrule
			Twitter   & 128  & 3 & 100 & 0.1  & 0.1  & 0.1  & 0.2 & 0.7 \\
                \midrule
			MUTAG   & 128  & 3 & 100 & 0.01  & 0.01  & 0.1  & 0.2 & 0.5 \\
                \midrule
			OGBG-Molbace   & 128  & 5 & 100 & 0.01  & 0.1  & 0.01  & 0.2 & 0.7 \\
                \midrule
			OGBG-Molbbbp   & 128  & 3 & 100 & 0.1  & 0.1  & 1  & 0.2 & 0.7 \\
                \midrule
			\bottomrule
	\end{tabular}}
	\label{parameters}
\end{table}

\textbf{Spurious-Motifs} \cite{ying2019gnnexplainer} is a synthetic graph classification dataset comprising three classes. Each graph within the dataset contains a motif $C$, which serves as the inherent explanation, and a base $S$ that exhibits a spurious correlation with the label. Specifically, in the training dataset, the motif is sampled from a uniform distribution, while the base is manually constructed according to $P(S)=b \times \mathbb{I}(S=C)+\frac{1-b}{2} \times \mathbb{I}(S \neq C)$, where $b$ is the bias rate \cite{wu2022discovering}. To introduce a distribution shift in the testing dataset, the motif and base are randomly connected to each other.

\textbf{Open Graph Benchmark (OGB)} \cite{hu2020open} is a large-scale dataset designed for graph machine learning tasks, covering a diverse range of domains from social networks to molecular graphs. We consider two OGB-Mol datasets for predicting molecular properties. The datasets are split by default, with each split containing a set of scaffolds distinct from the others.

\subsubsection{\textbf{Baselines}}
We validate the effectiveness of GRBE in two aspects: classification and rationalization performance. 
\begin{table*}[htbp]
	\centering
	\caption{The graph classification performance over synthetic and real-world datasets with different degrees of distribution shifts. The best results are in bold and the second-best results are underlined. Metric "Distance" indicates the distribution shift between training and testing datasets, with lower values indicating smaller distribution shifts.}
	\resizebox{0.975\textwidth}{3.25cm}{
		\begin{tabular}{c|ccc|ccc|c|cc}
			\toprule
			\midrule
			& Spmotif-0.9 & Spmotif-0.7 & Spmotif-0.5 & Graph-SST2 & Graph-SST5 & Twitter & MUTAG & OGBG-Molbace & OGBG-Molbbbp\\
			\midrule
			Metric   & \multicolumn{3}{c|}{ACC(\%)} & \multicolumn{3}{c|}{ACC(\%)} & \multicolumn{1}{c|}{ACC(\%)} & \multicolumn{2}{c}{AUC(\%)}\\
			\midrule
			Distance   & 11.466  & 11.183  & 10.280  & 5.9851  & 0.3127  & 0.0575 & 2.7303 & 0.2521  & 0.2703  \\
			\midrule
			GIN   & 38.570 $\pm$ 2.310 & 39.040 $\pm$ 1.620 & 39.870 $\pm$ 1.300 & 82.730 $\pm$ 0.770 & 36.660 $\pm$ 1.958 & 63.750 $\pm$ 2.795  & 91.653 $\pm$ 6.925 & \textbf{80.470 $\pm$ 1.720} & 65.840 $\pm$ 2.240 \\
			\midrule
			IRM   & 42.960 $\pm$ 9.424 & 46.040 $\pm$ 14.67 & 45.440 $\pm$ 7.143 & 80.120 $\pm$ 1.618 & 41.360 $\pm$ 1.799 & 63.500 $\pm$ 1.230 & 92.660 $\pm$ 2.721 & 74.240 $\pm$ 4.809 & 66.640 $\pm$ 2.876 \\
			IB-IRM & 41.660 $\pm$ 3.545 & \underline{54.140 $\pm$ 7.465} & 47.860 $\pm$ 7.422 & 80.620 $\pm$ 0.701 & 42.440 $\pm$ 0.841 & 61.260 $\pm$ 1.200 & 89.800 $\pm$ 5.443 & 76.700 $\pm$ 5.148 & 63.980 $\pm$ 2.644 \\
			V-REX & 41.780 $\pm$ 9.174 & 49.220 $\pm$ 14.14 & 43.960 $\pm$ 10.76 & 79.320 $\pm$ 2.487 & 40.760 $\pm$ 0.598 & 63.210 $\pm$ 1.570 & 91.040 $\pm$ 1.890 & 71.560 $\pm$ 6.317 & 65.940 $\pm$ 2.050 \\
			EIIL  & 40.720 $\pm$ 6.180 & 47.060 $\pm$ 5.076 & 43.160 $\pm$ 8.722 & 79.960 $\pm$ 1.847 & \underline{42.820 $\pm$ 0.931} & 62.760 $\pm$ 1.720 & 92.780 $\pm$ 2.064 & 73.280 $\pm$ 6.902 & 65.580 $\pm$ 2.167 \\
			\midrule
			GIB   & 46.190 $\pm$ 5.630 & 48.510 $\pm$ 5.760 & 54.360 $\pm$ 7.090 & \underline{82.990 $\pm$ 6.670} & 40.370 $\pm$ 6.298 & 63.298 $\pm$ 5.516  & 83.900 $\pm$ 6.400 & 69.100 $\pm$ 6.900 & 62.833 $\pm$ 4.640 \\
			GSAT  & 43.233 $\pm$ 2.802 & 48.950 $\pm$ 4.425 & 52.633 $\pm$ 1.115 & 82.167 $\pm$ 0.153 & 41.367 $\pm$ 3.742  & \underline{66.700 $\pm$ 3.637} & \underline{94.333 $\pm$ 1.002} & 72.023 $\pm$ 4.360 & 67.667 $\pm$ 1.365 \\
			DIR   & 37.610 $\pm$ 2.020 & 41.130 $\pm$ 2.620 & 45.490 $\pm$ 3.810  & 82.320 $\pm$ 0.850 & 41.007 $\pm$ 0.954 & 64.600 $\pm$ 3.637 & 87.533 $\pm$ 4.821 & \underline{79.333 $\pm$ 0.833} & 65.133 $\pm$ 0.451 \\
			GREA  & \underline{46.633 $\pm$ 8.695} & 50.813 $\pm$ 4.439 & \underline{54.717 $\pm$ 6.047} & 82.167 $\pm$ 1.501 & 40.120 $\pm$ 7.222 & 66.667 $\pm$ 7.217 & 93.853 $\pm$ 2.164 & 76.467 $\pm$ 4.131 & \textbf{68.733 $\pm$ 1.415} \\
			\midrule
			GRBE   & \textbf{61.533 $\pm$ 2.795} & \textbf{59.022 $\pm$ 7.273} & \textbf{62.956 $\pm$ 5.707} & \textbf{84.100 $\pm$ 0.587} & \textbf{43.846 $\pm$ 6.334} & \textbf{67.500 $\pm$ 6.847} & \textbf{94.798 $\pm$ 4.345} & 76.228 $\pm$ 2.154 & \underline{68.094 $\pm$ 1.670} \\
			\midrule
			Improve   & + 31.951\% & + 9.0173\% & + 15.055\% & + 1.3375\% & + 2.3961\% & + 1.1994\% & + 0.4929\% & - 5.5649\% & - 0.9384\% \\
			\midrule
			\bottomrule
	\end{tabular}}
	\label{classification_acc}
\end{table*}

\begin{table}[htbp]
	\centering
	\caption{The graph rationalization performance on synthetic and real-world datasets. The best results are in bold and the second-best results are underlined.}
	\resizebox{0.48\textwidth}{2.5cm}{
		\begin{tabular}{c|ccc|c}
			\toprule
			\midrule
			& Spmotif-0.9    & Spmotif-0.7    & Spmotif-0.5    & MUTAG\\
			\midrule
			GNNExplainer & 58.86 $\pm$ 1.93 & 62.25 $\pm$ 3.61 & 62.62 $\pm$ 1.35 & 61.98 $\pm$ 5.45\\
			PGExplainer  & \underline{72.34 $\pm$ 2.91} & 72.33 $\pm$ 9.18 & 69.54 $\pm$ 5.64 & 60.91 $\pm$ 17.1\\
			GraphMask    & 66.68 $\pm$ 6.96 & 73.06 $\pm$ 4.91 & 72.06 $\pm$ 5.58 & 62.23 $\pm$ 9.01\\
			\midrule
			GIB          & 47.29 $\pm$ 13.3 & 62.89 $\pm$ 15.5 & 57.29 $\pm$ 14.3 & 91.04 $\pm$ 6.59\\
			GSAT         & 69.40 $\pm$ 2.42 & \underline{78.26 $\pm$ 5.43} & \underline{78.23 $\pm$ 2.66} & \underline{99.16 $\pm$ 0.70} \\
			DIR          & 49.08 $\pm$ 3.66 & 77.68 $\pm$ 1.22 & 78.15 $\pm$ 1.32 & 87.53 $\pm$ 4.82\\
			GREA         & 57.91 $\pm$ 6.04 & 67.13 $\pm$ 3.76 & 66.94 $\pm$ 3.31 & 62.63 $\pm$ 4.23 \\
			\midrule
			GRBE          & \textbf{85.04 $\pm$ 2.34} & \textbf{85.47 $\pm$ 2.98} & \textbf{80.68 $\pm$ 2.72} & \textbf{99.86 $\pm$ 0.09}\\
			\midrule
			Improvement  & + 17.556\% & + 9.2128\% & + 3.1318\% & + 0.7059 \% \\
			\midrule
			\bottomrule
	\end{tabular}}
	\label{interpretation_AUC}
\end{table}%

\textbf{Graph-SST2} \cite{socher2013recursive} is a real-world sentiment graph dataset where each text instance is converted into a graph. To introduce distribution shifts, following the splits in \cite{wu2022discovering}, the average node degree in the training set is higher than that in the testing set. Each graph in SST2 is assigned to 2 classes according to the sentence sentiment. Additionally, Graph-SST5 and Graph-Twitter are two other sentiment graph datasets with 5 and 3 classes, respectively.

\textbf{MUTAG} \cite{rupp2012fast} is a widely used molecular dataset for explanation tasks. The graphs in MUTAG are labeled based on their mutagenic effects on bacteria. Using domain knowledge, specific chemical groups are assigned as ground truth explanations.

For classification performance, we compare GRBE with four main categories of related methods: (1) backbone GNN: GIN \cite{xu2018how}; (2) graph information bottleneck: GIB \cite{yu2021graph}, GSAT \cite{miao2022interpretable}; (3) graph invariant learning: IRM \cite{arjovsky2019invariant}, IB-IRM \cite{ahuja2021invariance}, VREX \cite{krueger2021out} and EIIL \cite{creager2021environment}; (4) graph rationalization: DIR \cite{wu2022discovering}, GREA \cite{liu2022graph}. The backbone GNN is considered to demonstrate the generalizability of GNN.

For rationalization performance, we compare GRBE with two different interpretable GNNs, which can be classified into the post-hoc and self-interpretable models. 
Post-hoc methods operate on a pre-trained GNN with fixed weights and identify a subgraph that dominates the prediction, which includes GNNExplainer \cite{ying2019gnnexplainer}, PGExplainer \cite{luo2020parameterized}, and GraphMask \cite{schlichtkrull2021interpreting}. In contrast, self-interpretable methods, such as GIB \cite{yu2021graph}, GSAT \cite{miao2022interpretable}, DIR \cite{wu2022discovering}, GREA \cite{liu2022graph}, integrate an interpretability module directly within the models, enabling them to find informative subgraphs and meanwhile making robust predictions.

\subsubsection{\textbf{Evaluation}}
We adopt two different metrics to evaluate the rationalization and classification performance of GRBE. To evaluate the rationalization performance, following the experimental setting of \cite{miao2022interpretable}, we employ AUC to evaluate the difference between the learned and ground-truth rationale subgraphs. Since not all datasets have ground-truth rationales, we report the rationalization performance on the Spmotif and MUTAG datasets. For classification performance, we use accuracy (ACC) as the evaluation metric for all the datasets, except for the OGBG dataset, where we use AUC following \cite{chen2022learning}. We report the mean results and standard deviation of the corresponding metrics across five runs. Meanwhile, to measure the difference between training and testing distributions, we adopt Jensen–Shannon divergence \cite{menendez1997jensen} as a distance metric:
\begin{equation}
	JS(p, q)=\frac{1}{2} KL\left(p \| \frac{p+q}{2}\right)+\frac{1}{2} KL\left(q \| \frac{p+q}{2}\right)
\end{equation}
where $KL(p \| q)=\int_{x} p(x) \log \frac{p(x)}{q(x)} dx$ is the Kullback-Leibler Divergence, $p$ and $q$ are two distributions.

\subsection{Implementation Details}
We adopt the GIN as the backbone GNN in our model, and the number of GNN layers is selected from \{3, 5\} and the number of hidden layer sizes is chosen from \{64,128,256\}. We do not do exhaustive hyperparameter tuning for the loss. For the coefficients $\alpha, \beta, \gamma$, we tune their values from \{0.001, 0.01, 0.1, 1, 5,10, 100\}, and from \{0.1, 0.2, 0.3, 0.4, 0.5, 0.6, 0.7\} for the augmentation ratio $r$ and sparsity level $r_s$. The number of epochs is set as 30, 100, or 200. The model is trained via an Adam optimizer with a default learning rate of 0.001. We show the detailed parameter combinations in the Table \ref{parameters}.
We implement the GRBE method based on Pytorch 1.13.0 and PyG (PyTorch Geometric 2.0.2) graph deep learning library. The experiments are conducted on a server with NVIDIA GeForce RTX 3090 GPUs and a server with NVIDIA A100 Tensor Core GPU.

\begin{figure}[htbp]
	\centering
 	\subfloat{
		\includegraphics[width=0.225\textwidth]{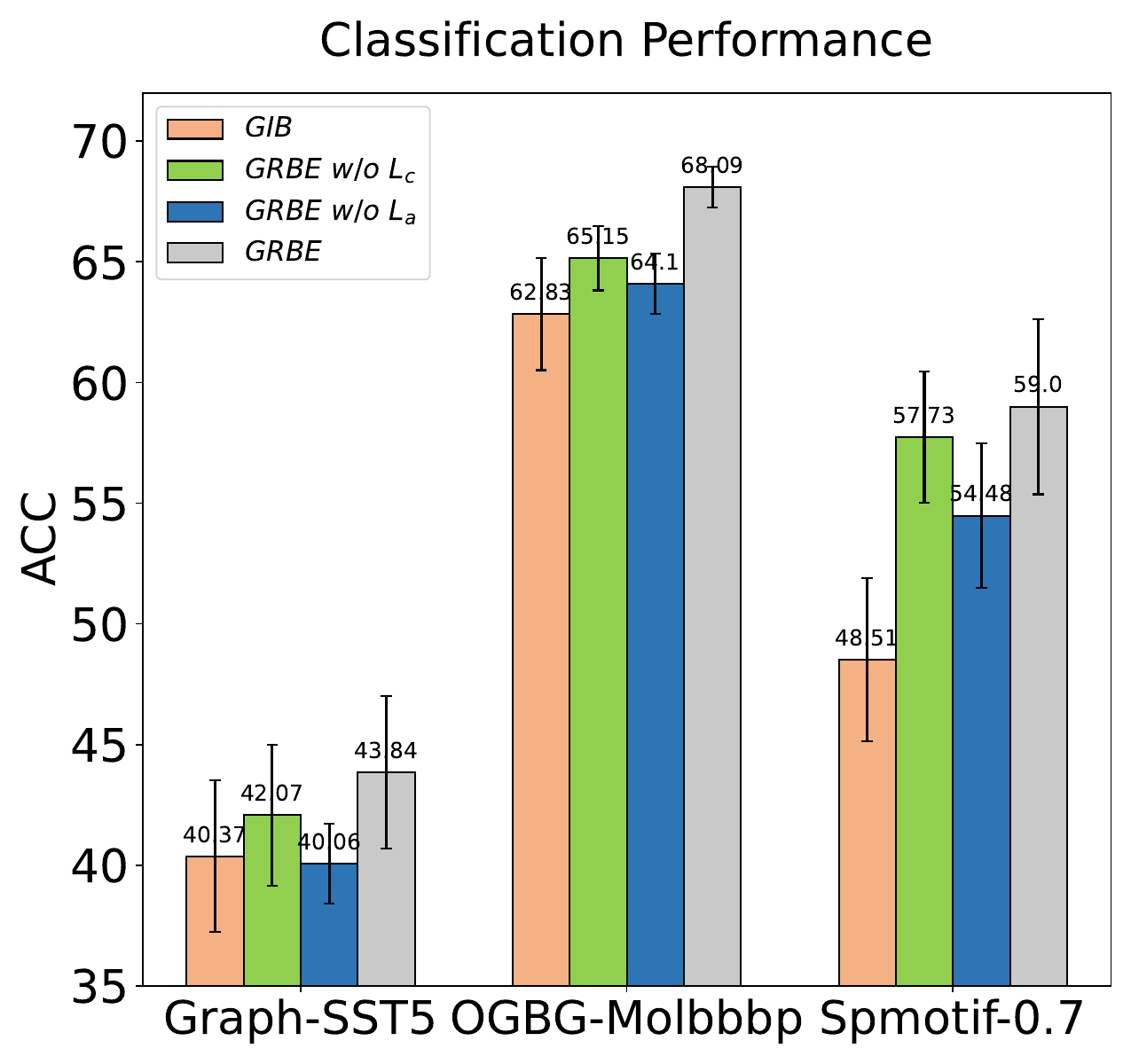}
		\label{ablation_classification}
	}
	\subfloat{
		\includegraphics[width=0.225\textwidth]{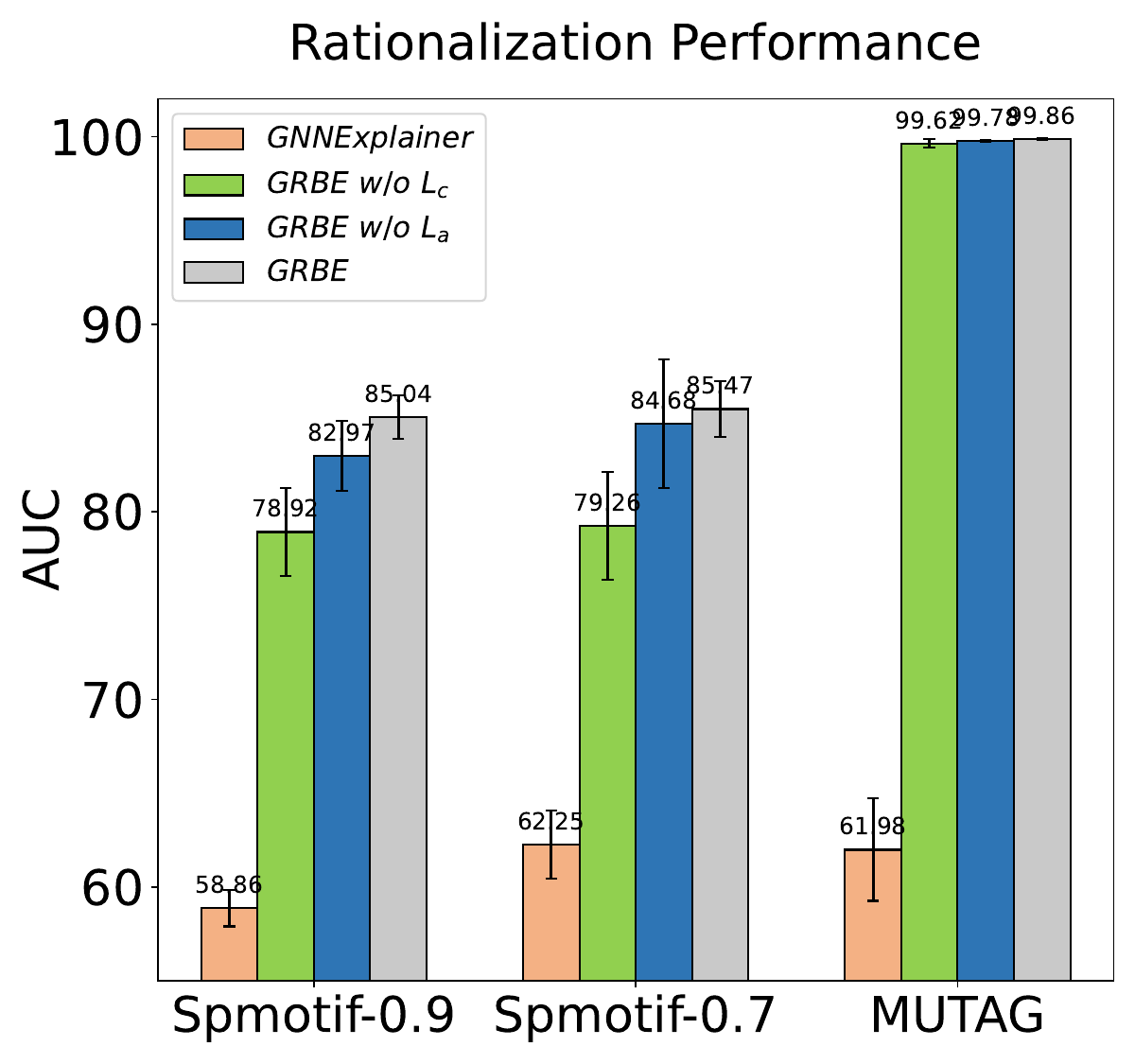}
		\label{ablation_rationalization}
	}
	\caption{Ablation studies. We report the mean - 0.5*std of the rationalization and classification performance of the different variants of GRBE.}
	\label{ablation}
\end{figure}

\begin{figure*}[htbp]
	\centering
	\subfloat{
		\includegraphics[width=0.23\textwidth]{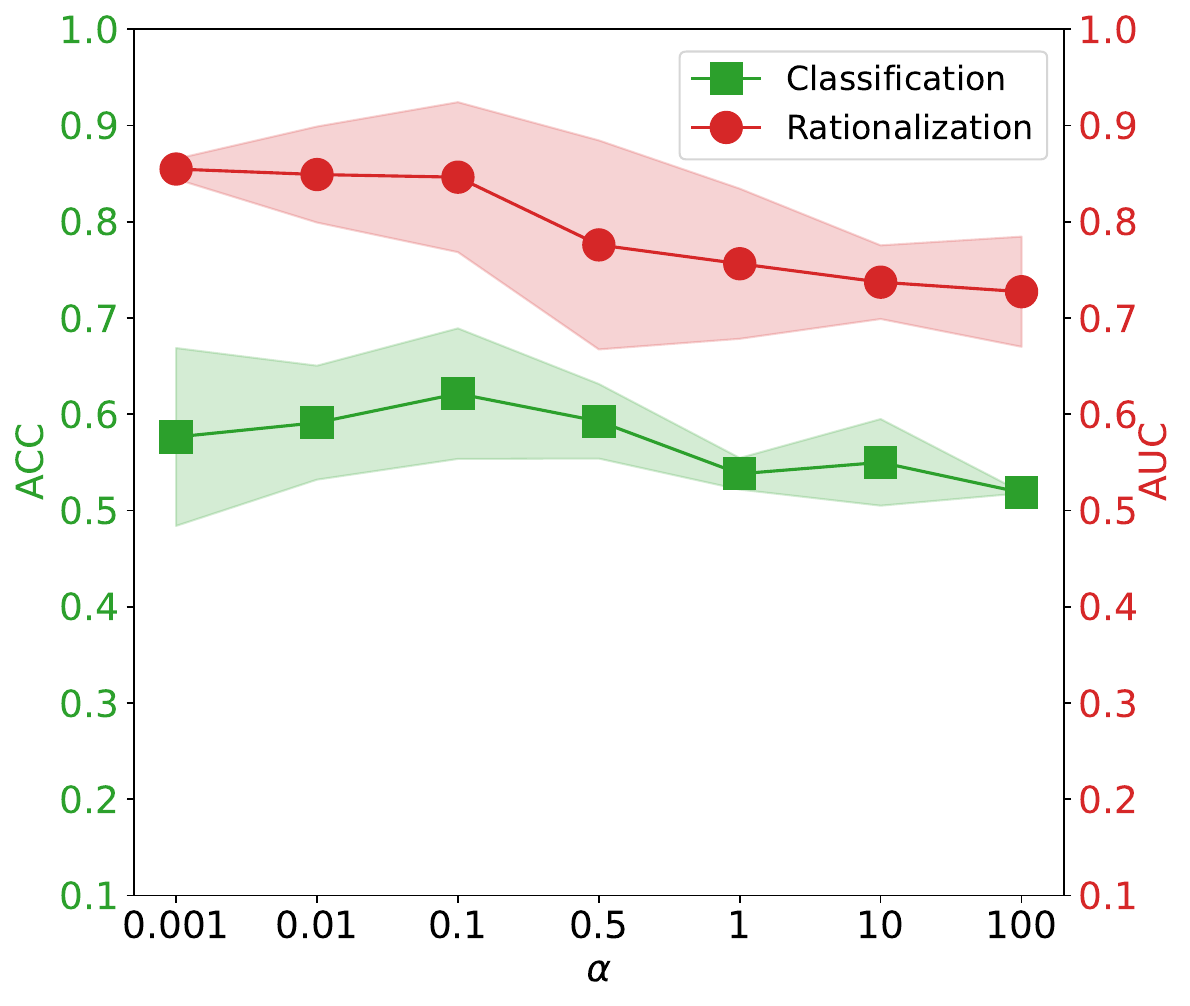}
		\label{sensitive_aug}
	}
	\subfloat{
		\includegraphics[width=0.23\textwidth]{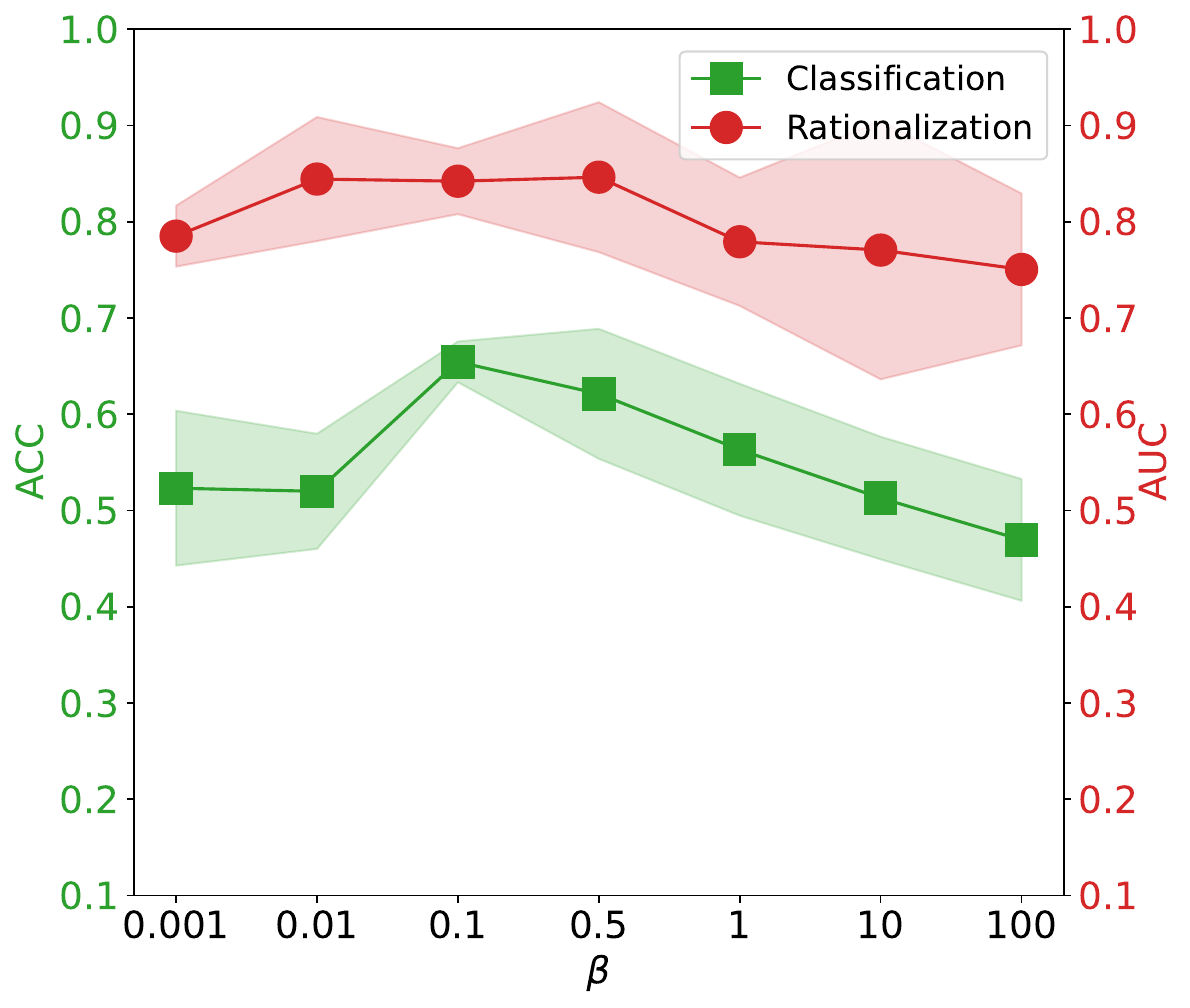}
		\label{sensitive_con}
	}
	\subfloat{
		\includegraphics[width=0.23\textwidth]{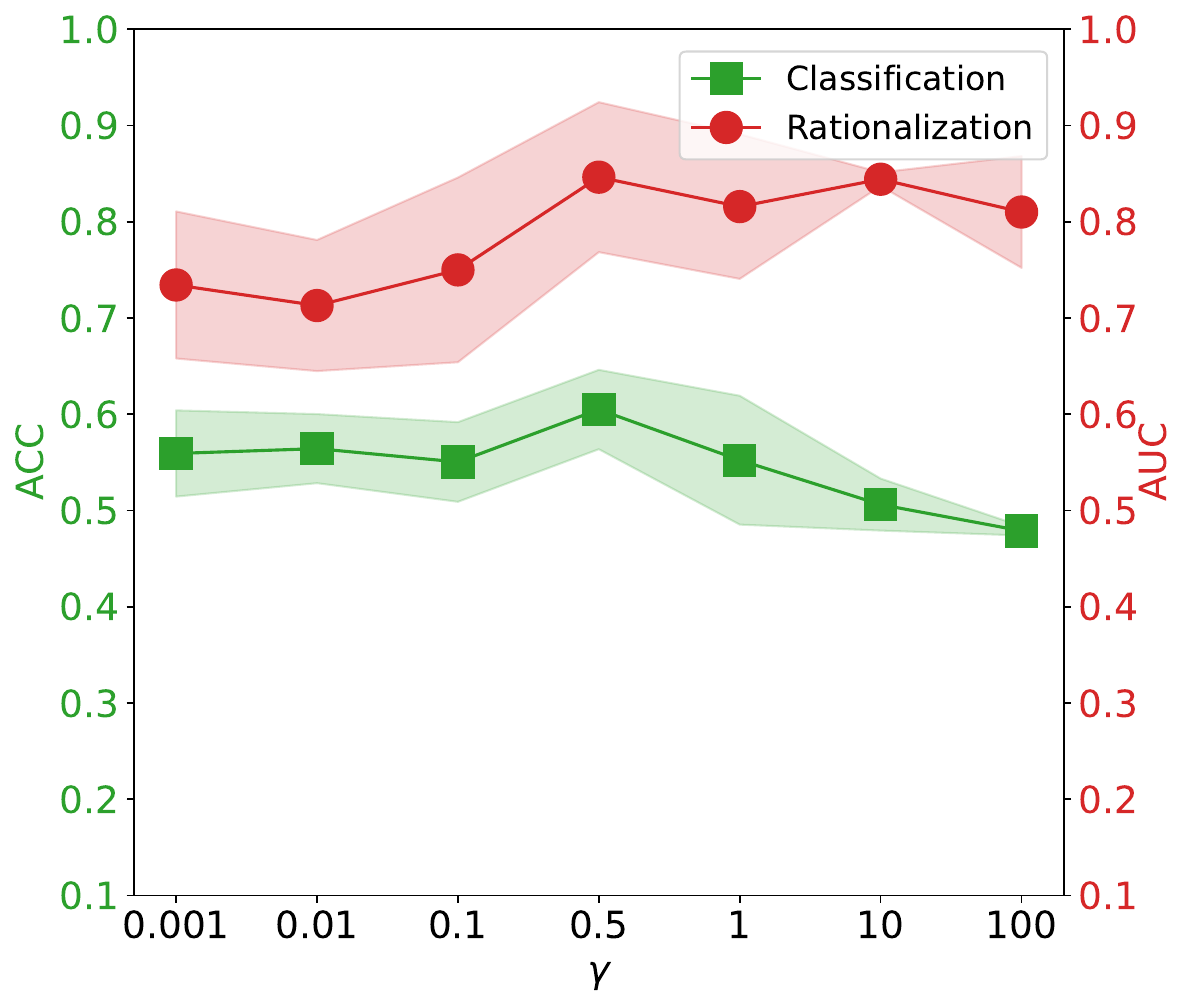}
		\label{sensitive_spa}
	}\\
 	\subfloat{
		\includegraphics[width=0.23\textwidth]{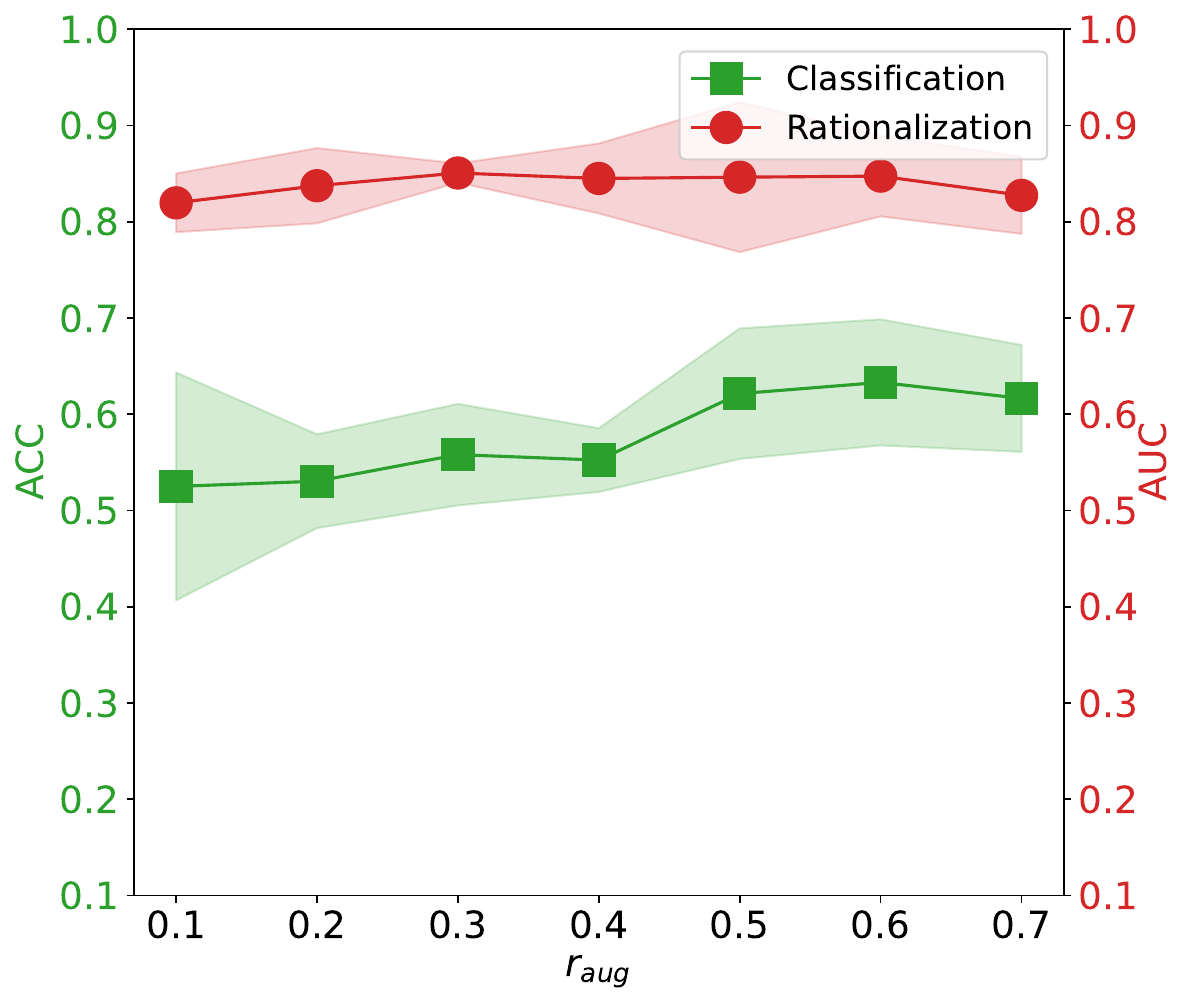}
		\label{aug_r}
	}
	\subfloat{
		\includegraphics[width=0.23\textwidth]{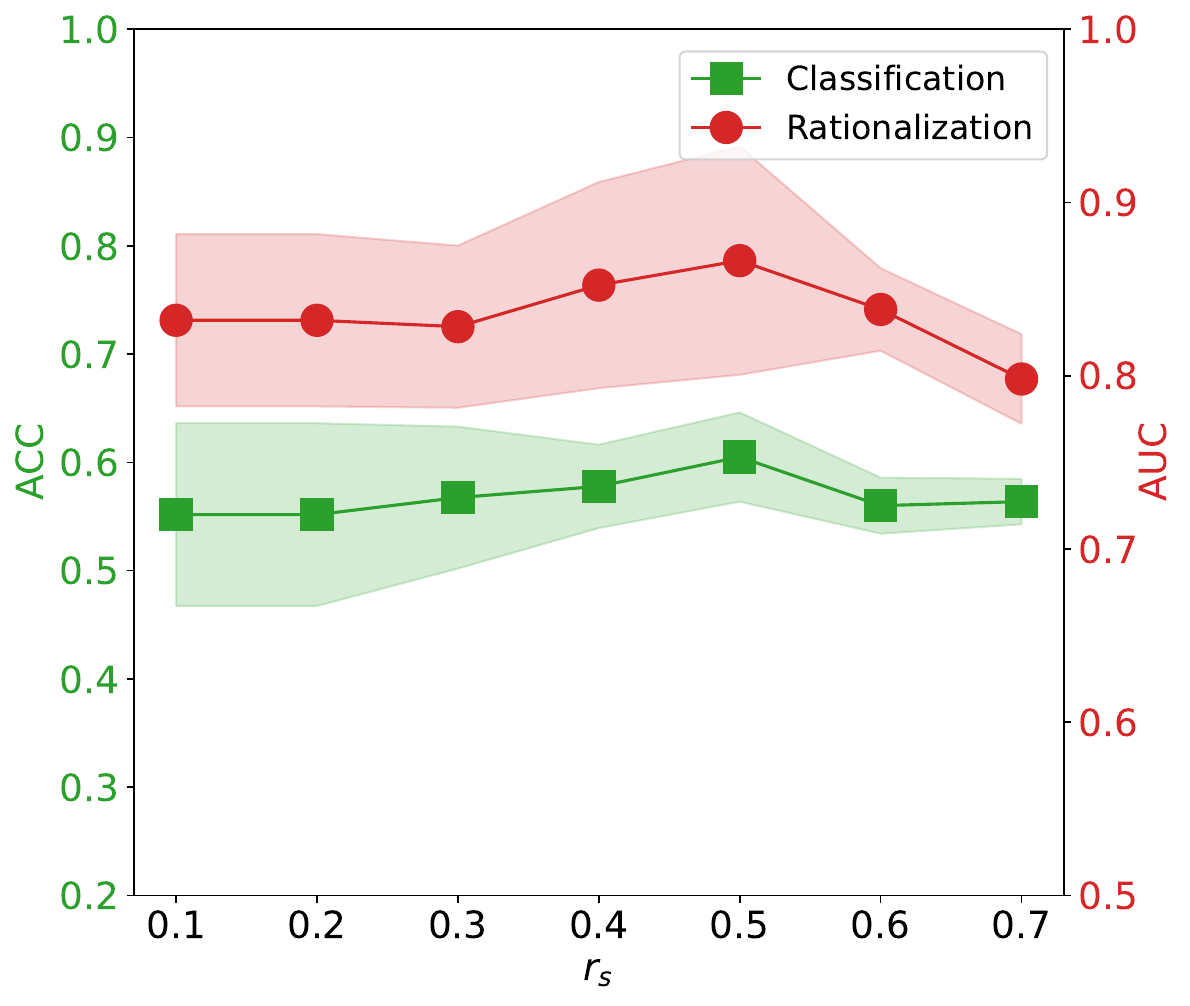}
		\label{sensitive_r}
	}
	\caption{Hyper-parameter sensitive analysis. We report the mean - std of classification and rationalization performance under five critical hyper-parameters $\alpha, \beta, \gamma, r_{aug}, r_s$ on the Spmotif-0.7 dataset.}
	\label{sensitive}
\end{figure*}

\subsection{Effectiveness Results}
\textbf{Classification Performance.} We report the classification performance and the distance metric in Table \ref{classification_acc}. We adopt the JS divergence as a distance to measure the difference between the training and testing distributions, for ease of presentation, we give the distance result divided by 1e3.  Table \ref{classification_acc} leads to the following observations. Firstly, we can see that GRBE outperforms other methods by substantial margins on the Spmotif datasets, ranging from $+9.0173\%$ to $+31.951\%$. Additionally, GRBE surpasses its competitors on the Graph-SST2, Graph-SST5, Twitter, and MUTAG datasets. However, GRBE is slightly inferior to the baselines on the OGBG-Molbace and OGBG-Molbbbp datasets, which can be attributed to the smaller distribution shifts as indicated by the distance metric.
As the distance increases, indicating greater distribution shifts between the training and testing sets, the performance improvement of the GRBE method becomes more pronounced. This suggests that the augmented diversified graph can effectively mitigate the impact of distribution shifts on model predictions.

\textbf{Rationalization Performance.} We report the rationalization performance in Table \ref{interpretation_AUC}. As illustrated in Table \ref{interpretation_AUC}, GRBE achieves the highest AUC scores across all datasets and significantly surpasses other baselines on the Spmotif dataset. Specifically, the significant improvements over other methods, ranging from +0.7038\%
to +17.5670\%, effectively proving that GRBE can guide the attention to focus on the invariant rationale subgraphs. Meanwhile, as the bias $b$ increases, the AUC of the DIR and GREA show a significant decrease, indicating that these methods do not effectively capture invariance, while GRBE can maintain stable results. Overall, GRBE's consistent outperformance validates its effectiveness in achieving more precise subgraph division.

\subsection{Ablation Studies}
In this section, we respectively adopt "$GRBE\ w/o\ L_c$" and "$GRBE\ w/o\ L_a$" to represent the GRBE method without PRSE and EDA modules. We compare them with standard GRBE and two baselines to validate their effectiveness.
Fig. \ref{ablation} firstly shows the classification ACC performance, we observe that the standard GRBE consistently achieves the highest accuracy across all datasets, compared with its variants and the backbone GIN. Meanwhile, Fig. \ref{ablation} highlights the AUC performance for rationalization. Here, "$GRBE\ w/o\ L_c$" performs significantly worse than GRBE on the Spmotif datasets but still surpasses the GNNExplainer, demonstrating the superior interpretability provided by PRSE. On the MUTAG dataset, while the AUC of both variants is lower than that of GRBE, the difference is minimal. This is likely because MUTAG is a relatively simple dataset where existing GNNs already achieve high accuracy, limiting the AUC improvement of the variants. Overall, GRBE outperforms "$GRBE\ w/o\ L_c$" and "$GRBE\ w/o\ L_a$" in both classification ACC and rationalization AUC, which demonstrates that both the "$GRBE\ w/o\ L_c$" and "$GRBE\ w/o\ L_a$" modules are effective.

\subsection{Hyper-Parameter Sensitive Analysis}
To validate the hyper-parameter sensitivity of GRBE, we conduct experiments on the Spmotif-0.7 dataset to validate the sensitivity of five critical hyper-parameter $\alpha, \beta, \gamma, r_{aug}, r_s$. For the loss coefficients $\alpha, \beta, \gamma$, we tune their values from 1e-3 to 1e2, and from 0.1 to 0.7 for the augmentation ratio $r_{aug}$ and sparsity level $r_s$. From the classification ACC and rationalization AUC performance depicted in Fig. \ref{sensitive}, we can see that our GRBE method exhibits overall robustness to changes in hyper-parameters. Across all subplots, the performance maintains relatively stable performance with minor variations as the hyper-parameters are adjusted, and both the performance would be better when the loss coefficients are 0.1 to 1 and the sparsity level ranges from 0.3 to 0.5.

\begin{table*}[htbp]
	\centering
	\caption{The number of environment subgraph categories learned by GREA and GRBE on the synthetic and real datasets.}
	\resizebox{0.975\textwidth}{0.875cm}{
		\begin{tabular}{c|c|c|c|c|c|c|c|c|c}
			\toprule
                \midrule
			& Spmotif-0.9 & Spmotif-0.7 & Spmotif-0.5 & Graph-SST2 & Graph-SST5 & Twitter & MUTAG & OGBG-Molbace & OGBG-Molbbbp\\
			\midrule
			GREA   &  5& 2&   4&  22&  54& 11& 157 & 6 & 33 \\
			\midrule
			GRBE   & \textbf{34}& \textbf{68}& \textbf{36}&  \textbf{34}&  \textbf{81}& \textbf{31}& \textbf{323}& \textbf{25} & \textbf{156}\\
                \midrule
			\bottomrule
	\end{tabular}}
	\label{environment_counter}
\end{table*}

\begin{figure}[htbp]
	\centering
	\subfloat{
		\includegraphics[width=0.21\textwidth]{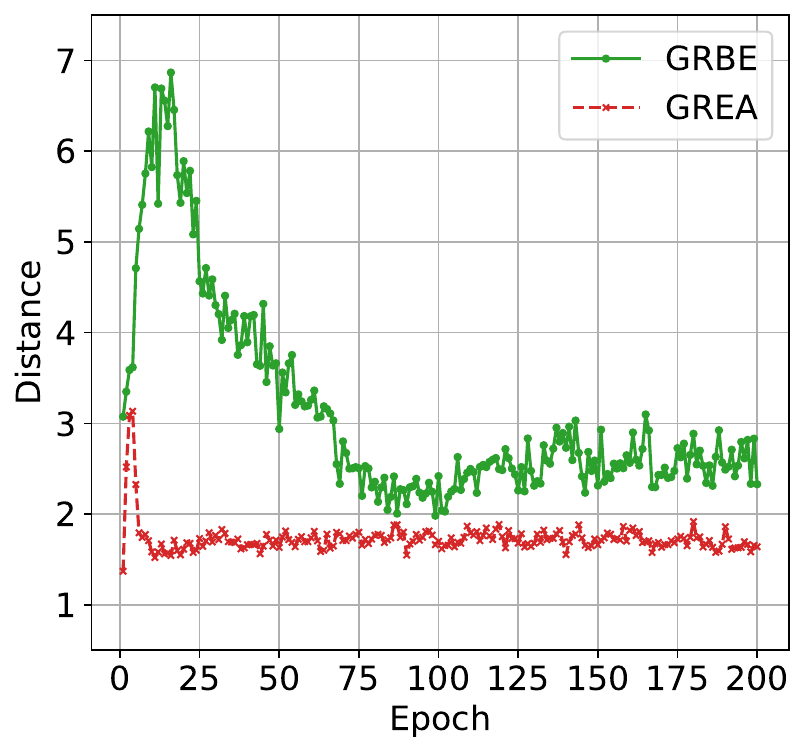}
		\label{JS_distance}
	}
	\subfloat{
		\includegraphics[width=0.235\textwidth]{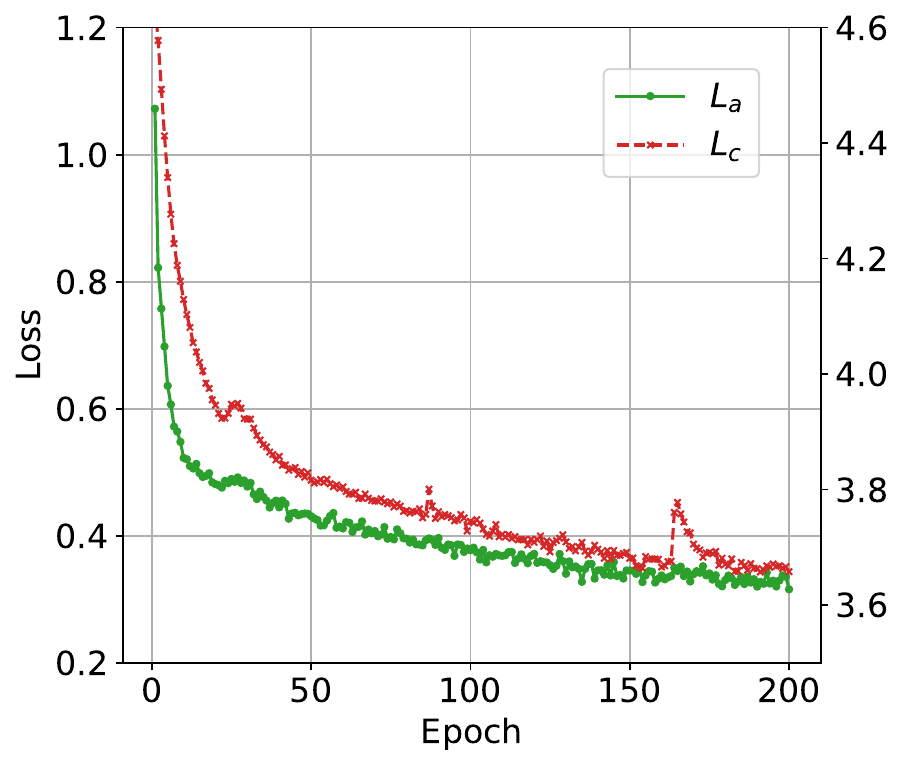}
		\label{GRBE_cluster}
	}
	\caption{The distance between the augmented distributions obtained by GRBE and GREA and the original distribution under each epoch, and the augmentation loss $L_a$ along with the contrastive loss $L_c$ of GRBE.}
	\label{Case_Study1}
\end{figure}

\subsection{Case Study}
In this subsection, we first conduct case studies on the Spmotif-0.9 dataset to validate the capability of GRBE in producing sufficiently diverse samples. We respectively utilize GRBE and GREA to generate an equivalent number of augmented samples as those in the training set. Then, we calculate the distance between their augmented distributions and the original distribution under each epoch, as depicted in Fig. \ref{Case_Study1}. We can see that, in the initial stages, GRBE exhibits a rapid increase in sample diversity due to potentially inaccurate subgraph divisions. Subsequently, the diversity of augmented graphs stabilizes and remains higher than what the GREA algorithm achieves. Also, we report the augmentation loss $L_a$ and the contrastive loss $L_c$ in Fig. \ref{Case_Study1}. The loss curves exhibit similar decreasing trends, demonstrating the mutual enhancement of the PRSE and EDA modules.

Secondly, we show the number of environment subgraph categories learned by GREA and GRBE in Table \ref{environment_counter}, we can see that GRBE can obtain more categories of environment subgraphs than the GREA algorithm on all datasets. When the distribution shift between the training and testing set is large, such as on the Spmotif dataset, the diversified augmented environment subgraphs will greatly improve the classification performance. However, in datasets with minimal distribution shifts, like OGBG-molbace and OGBG-molbbbp, increasing the diversity of augmented environment subgraphs may amplify the distribution shift between training and testing data, leading to poorer classification performance. 

\begin{figure}[htbp]
	\centering
	\subfloat{
		\includegraphics[width=0.475\linewidth]{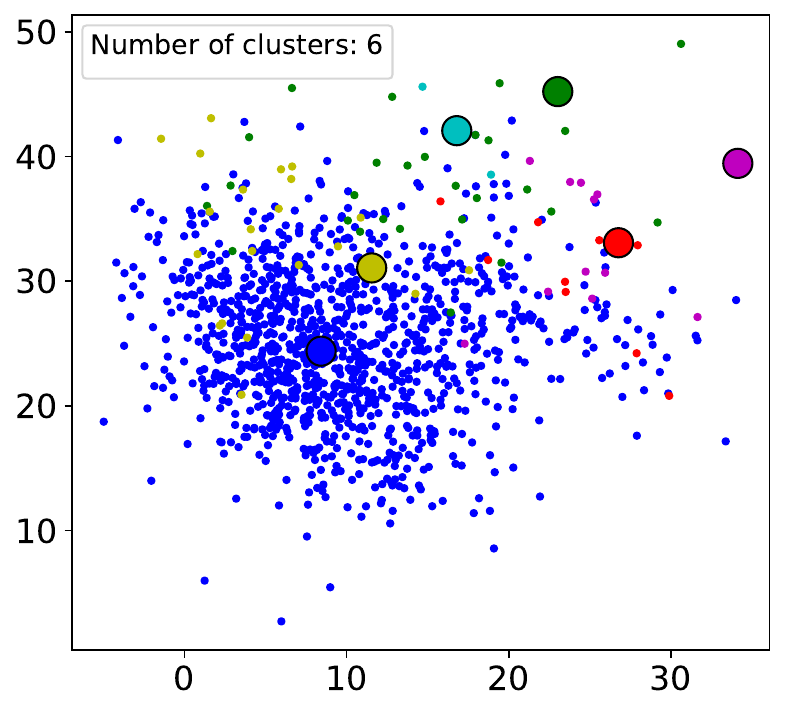}
		\label{Spmotif_cluster}
	}
	\subfloat{
		\includegraphics[width=0.46\linewidth]{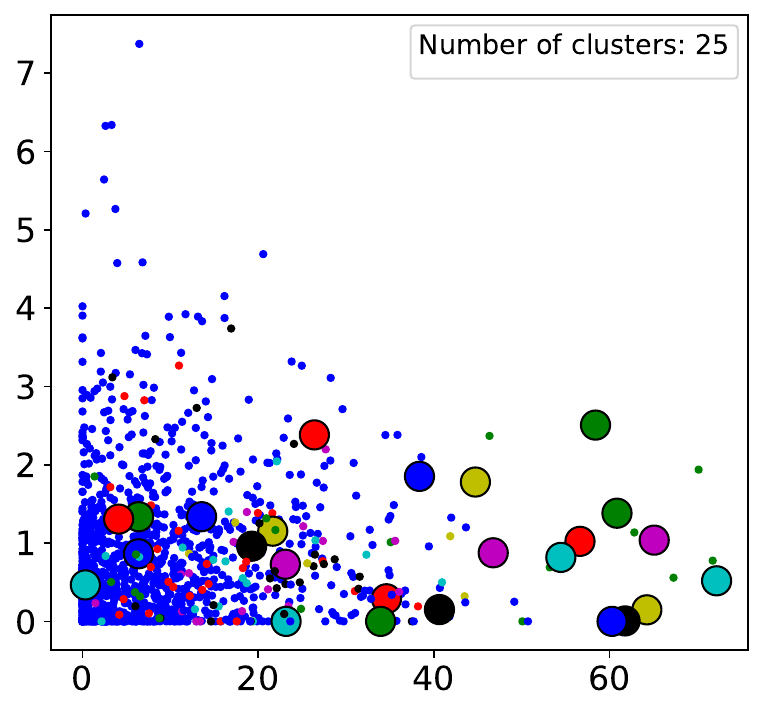}
		\label{Molbace_cluster}
	}
	\caption{The unsupervised clustering results of the environment subgraph representations on the OGBG-Molbace dataset. (a) The representations are learned by the existing GREA method. (b) The representations are obtained using our proposed GRBE method.}
	\label{Case_Study2}
\end{figure}

Meanwhile, Fig. \ref{Case_Study2} presents clustering results of environment subgraphs augmented by GREA and GRBE on the OGBG-Molbace datasets. In contrast to the 6 clusters generated by the GREA on the OGBG-Molbace dataset, GRBE produces 25 clusters, indicating its ability to generate significantly more diverse environment subgraphs. Lastly, the above observations show that our method can effectively generate a more diverse training distribution.

\begin{figure*}[htbp]
	\centering
        \includegraphics[width=1\textwidth]{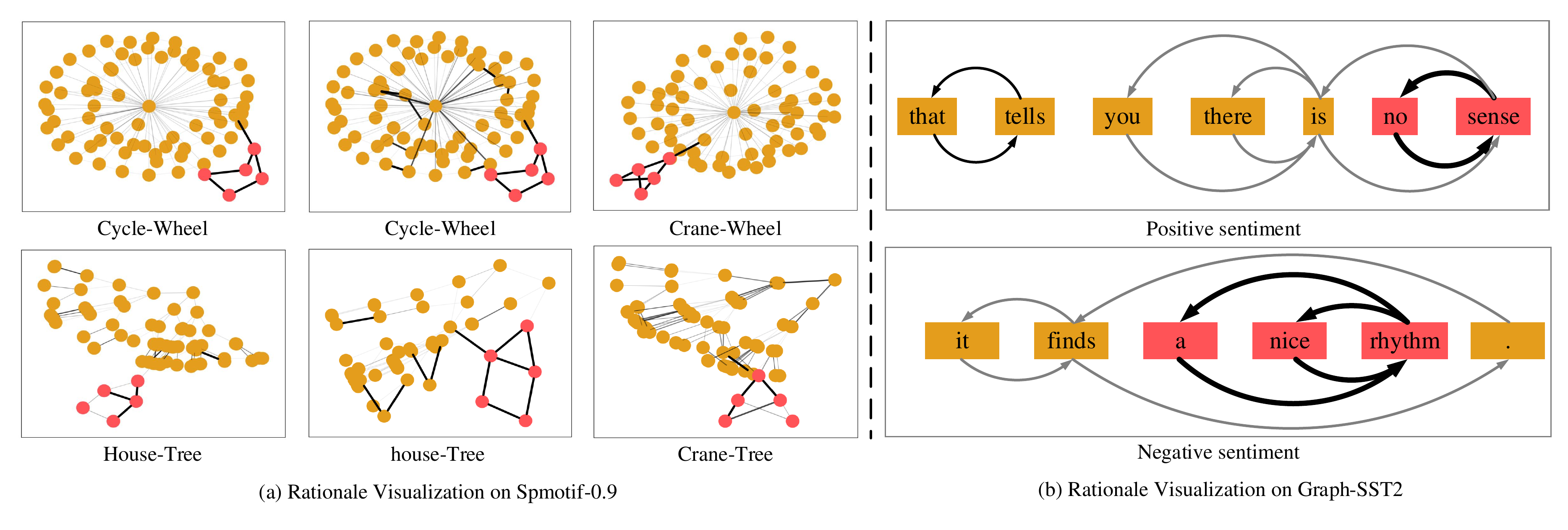}
	\caption{Visualization of the rationale subgraphs extracted by GRBE on the Spmotif-0.9 and Graph-SST2 datasets. The ground-truth rationales are colored in red. The rationale subgraphs identified by GRBE are highlighted with black lines, with darker colors indicating higher edge weights.}
	\label{visualization}
\end{figure*}

\subsection{Rationale Subgraph Visualization}
To verify GRBE's ability to find ground-truth rationale subgraphs, we provide visualizations of the identified rationale subgraphs on the synthetic Spmotif-0.9 and real-word Graph-SST2 datasets. as shown in Fig. \ref{visualization}. In the Spmotif-0.9 dataset, each graph consists of a motif as the rationale subgraph and a base as the environment subgraph. In Fig. \ref{visualization} (a), regardless of how the environment subgraph changes, we can see that the rationale subgraphs learned by GRBE contain most of the edges within the ground-truth rationale subgraphs and only a few spurious edges from environment subgraphs. For the Graph-SST2 dataset that has no ground-truth rationale subgraphs, in Fig. \ref{visualization} (b), GRBE identifies "no sense" for the negative sentiment and "a nice rhythm" for the positive sentiment, aligning with human intuition. These observations indicate that our proposed GRBE effectively identifies precise rationale subgraphs from the input graphs.

\section{Conclusion}
In this work, we proposed a GRBE method, which improves graph rationalization under OOD settings. Specifically, GRBE first utilizes a PRSE module to guide the rationale subgraph generator achieving more precise subgraph division. Then, an EDA strategy is designed to explore unknown environment subgraphs and produce an augmented distribution with greater coverage. Extensive experiments and analyses show that our GRBE significantly outperforms state-of-the-art methods in most synthetic and real-world datasets, demonstrating its effectiveness. In the future, how to select the best individuals from a large number of augmented samples to better adapt to the generalization task on the graph is also an interesting direction.

\section*{Acknowledgments}
This work is supported by the National Science and Technology Major Project of China (2021ZD0111801) and the National Natural Science Foundation of China (under grant 62376087).

\bibliographystyle{IEEEtran}
\bibliography{reference}

\begin{thebibliography}{10}
\providecommand{\url}[1]{#1}
\csname url@samestyle\endcsname
\providecommand{\newblock}{\relax}
\providecommand{\bibinfo}[2]{#2}
\providecommand{\BIBentrySTDinterwordspacing}{\spaceskip=0pt\relax}
\providecommand{\BIBentryALTinterwordstretchfactor}{4}
\providecommand{\BIBentryALTinterwordspacing}{\spaceskip=\fontdimen2\font plus
\BIBentryALTinterwordstretchfactor\fontdimen3\font minus \fontdimen4\font\relax}
\providecommand{\BIBforeignlanguage}[2]{{%
\expandafter\ifx\csname l@#1\endcsname\relax
\typeout{** WARNING: IEEEtran.bst: No hyphenation pattern has been}%
\typeout{** loaded for the language `#1'. Using the pattern for}%
\typeout{** the default language instead.}%
\else
\language=\csname l@#1\endcsname
\fi
#2}}
\providecommand{\BIBdecl}{\relax}
\BIBdecl

\bibitem{liu2022pre}
\BIBentryALTinterwordspacing
S.~Liu, H.~Wang, W.~Liu, J.~Lasenby, H.~Guo, and J.~Tang, ``Pre-training molecular graph representation with 3d geometry,'' in \emph{Proceedings of the 10th International Conference on Learning Representations}, 2022. [Online]. Available: \url{https://openreview.net/forum?id=xQUe1pOKPam}
\BIBentrySTDinterwordspacing

\bibitem{li2022out}
H.~Li, X.~Wang, Z.~Zhang, and W.~Zhu, ``Out-of-distribution generalization on graphs: A survey,'' \emph{arXiv preprint arXiv:2202.07987}, 2022.

\bibitem{miao2022interpretable}
S.~Miao, M.~Liu, and P.~Li, ``Interpretable and generalizable graph learning via stochastic attention mechanism,'' in \emph{Proceedings of the 39th International Conference on Machine Learning}, 2022, pp. 15\,524--15\,543.

\bibitem{yang2023individual}
L.~Yang, J.~Zheng, H.~Wang, Z.~Liu, Z.~Huang, S.~Hong, W.~Zhang, and B.~Cui, ``Individual and structural graph information bottlenecks for out-of-distribution generalization,'' \emph{IEEE Transactions on Knowledge and Data Engineering}, pp. 682--693, 2023.

\bibitem{jia2024graph}
T.~Jia, H.~Li, C.~Yang, T.~Tao, and C.~Shi, ``Graph invariant learning with subgraph co-mixup for out-of-distribution generalization,'' in \emph{Proceedings of the 38th AAAI Conference on Artificial Intelligence}, 2024, pp. 8562--8570.

\bibitem{chen2022learning}
Y.~Chen, Y.~Zhang, Y.~Bian, H.~Yang, M.~Kaili, B.~Xie, T.~Liu, B.~Han, and J.~Cheng, ``Learning causally invariant representations for out-of-distribution generalization on graphs,'' in \emph{Proceedings of the 36th Conference on Neural Information Processing Systems}, 2022, pp. 22\,131--22\,148.

\bibitem{gui2024joint}
S.~Gui, M.~Liu, X.~Li, Y.~Luo, and S.~Ji, ``Joint learning of label and environment causal independence for graph out-of-distribution generalization,'' in \emph{Proceedings of the 38th Conference on Neural Information Processing Systems}, 2024, pp. 3945--3978.

\bibitem{sui2022causal}
Y.~Sui, X.~Wang, J.~Wu, M.~Lin, X.~He, and T.-S. Chua, ``Causal attention for interpretable and generalizable graph classification,'' in \emph{Proceedings of the 28th ACM SIGKDD Conference on Knowledge Discovery and Data Mining}, 2022, pp. 1696--1705.

\bibitem{li2022ood}
H.~Li, X.~Wang, Z.~Zhang, and W.~Zhu, ``Ood-gnn: Out-of-distribution generalized graph neural network,'' \emph{IEEE Transactions on Knowledge and Data Engineering}, vol.~35, no.~7, pp. 7328--7340, 2022.

\bibitem{yang2021learning}
S.~Yang, K.~Yu, F.~Cao, L.~Liu, H.~Wang, and J.~Li, ``Learning causal representations for robust domain adaptation,'' \emph{IEEE Transactions on Knowledge and Data Engineering}, vol.~35, no.~3, pp. 2750--2764, 2021.

\bibitem{sun2024dive}
X.~Sun, L.~Wang, Q.~Liu, S.~Wu, Z.~Wang, and L.~Wang, ``Dive: Subgraph disagreement for graph out-of-distribution generalization,'' in \emph{Proceedings of the 30th ACM SIGKDD Conference on Knowledge Discovery and Data Mining}, 2024, pp. 2794--2805.

\bibitem{wu2022discovering}
\BIBentryALTinterwordspacing
Y.~Wu, X.~Wang, A.~Zhang, X.~He, and T.-S. Chua, ``Discovering invariant rationales for graph neural networks,'' in \emph{Proceedings of the 10th International Conference on Learning Representations}, 2022. [Online]. Available: \url{https://openreview.net/forum?id=hGXij5rfiHw}
\BIBentrySTDinterwordspacing

\bibitem{liu2022graph}
G.~Liu, T.~Zhao, J.~Xu, T.~Luo, and M.~Jiang, ``Graph rationalization with environment-based augmentations,'' in \emph{Proceedings of the 28th ACM SIGKDD Conference on Knowledge Discovery and Data Mining}, 2022, pp. 1069--1078.

\bibitem{sui2022adversarial}
Y.~Sui, X.~Wang, J.~Wu, A.~Zhang, and X.~He, ``Adversarial causal augmentation for graph covariate shift,'' \emph{arXiv preprint arXiv:2211.02843}, 2022.

\bibitem{comaniciu2002mean}
D.~Comaniciu and P.~Meer, ``Mean shift: A robust approach toward feature space analysis,'' \emph{IEEE Transactions on Pattern Analysis and Machine Intelligence}, vol.~24, no.~5, pp. 603--619, 2002.

\bibitem{ying2019gnnexplainer}
Z.~Ying, D.~Bourgeois, J.~You, M.~Zitnik, and J.~Leskovec, ``Gnnexplainer: Generating explanations for graph neural networks,'' in \emph{Proceedings of the 33rd Conference on Neural Information Processing Systems}, 2019, pp. 9244--9255.

\bibitem{chen2024does}
Y.~Chen, Y.~Bian, K.~Zhou, B.~Xie, B.~Han, and J.~Cheng, ``Does invariant graph learning via environment augmentation learn invariance?'' vol.~36, 2024.

\bibitem{yu2021graph}
\BIBentryALTinterwordspacing
J.~Yu, T.~Xu, Y.~Rong, Y.~Bian, J.~Huang, and R.~He, ``Graph information bottleneck for subgraph recognition,'' in \emph{Proceedings of the 9th International Conference on Learning Representations}, 2021. [Online]. Available: \url{https://openreview.net/forum?id=bM4Iqfg8M2k}
\BIBentrySTDinterwordspacing

\bibitem{yu2022improving}
J.~Yu, J.~Cao, and R.~He, ``Improving subgraph recognition with variational graph information bottleneck,'' in \emph{Proceedings of the 32nd IEEE/CVF Conference on Computer Vision and Pattern Recognition}, 2022, pp. 19\,396--19\,405.

\bibitem{yang2024individual}
L.~Yang, J.~Zheng, H.~Wang, Z.~Liu, Z.~Huang, S.~Hong, W.~Zhang, and B.~Cui, ``Individual and structural graph information bottlenecks for out-of-distribution generalization,'' \emph{IEEE Transactions on Knowledge and Data Engineering}, vol.~36, no.~2, pp. 682--693, 2024.

\bibitem{arjovsky2019invariant}
M.~Arjovsky, L.~Bottou, I.~Gulrajani, and D.~Lopez-Paz, ``Invariant risk minimization,'' \emph{arXiv preprint arXiv:1907.02893}, 2019.

\bibitem{ahuja2021invariance}
K.~Ahuja, E.~Caballero, D.~Zhang, J.-C. Gagnon-Audet, Y.~Bengio, I.~Mitliagkas, and I.~Rish, ``Invariance principle meets information bottleneck for out-of-distribution generalization,'' in \emph{Proceedings of the 35th Conference on Neural Information Processing Systems}, 2021, pp. 3438--3450.

\bibitem{krueger2021out}
D.~Krueger, E.~Caballero, J.-H. Jacobsen, A.~Zhang, J.~Binas, D.~Zhang, R.~Le~Priol, and A.~Courville, ``Out-of-distribution generalization via risk extrapolation (rex),'' in \emph{Proceedings of the 38th International Conference on Machine Learning}, 2021, pp. 5815--5826.

\bibitem{yang2022learning}
N.~Yang, K.~Zeng, Q.~Wu, X.~Jia, and J.~Yan, ``Learning substructure invariance for out-of-distribution molecular representations,'' \emph{Advances in Neural Information Processing Systems}, vol.~35, pp. 12\,964--12\,978, 2022.

\bibitem{li2022learning}
H.~Li, Z.~Zhang, X.~Wang, and W.~Zhu, ``Learning invariant graph representations for out-of-distribution generalization,'' in \emph{Proceedings of the 36th Conference on Neural Information Processing Systems}, 2022, pp. 11\,828--11\,841.

\bibitem{liu2023flood}
Y.~Liu, X.~Ao, F.~Feng, Y.~Ma, K.~Li, T.-S. Chua, and Q.~He, ``Flood: A flexible invariant learning framework for out-of-distribution generalization on graphs,'' in \emph{Proceedings of the 29th ACM SIGKDD Conference on Knowledge Discovery and Data Mining}, 2023, pp. 1548--1558.

\bibitem{fan2022debiasing}
S.~Fan, X.~Wang, Y.~Mo, C.~Shi, and J.~Tang, ``Debiasing graph neural networks via learning disentangled causal substructure,'' \emph{Advances in Neural Information Processing Systems}, vol.~35, pp. 24\,934--24\,946, 2022.

\bibitem{wang2024advancing}
R.~Wang, H.~Dai, C.~Yang, L.~Song, and C.~Shi, ``Advancing molecule invariant representation via privileged substructure identification,'' in \emph{Proceedings of the 30th ACM SIGKDD Conference on Knowledge Discovery and Data Mining}, 2024, pp. 3188--3199.

\bibitem{yue2024learning}
\BIBentryALTinterwordspacing
L.~Yue, Q.~Liu, Y.~Liu, W.~Gao, and C.~Song, ``Learning from shortcut: A shortcut-guided approach for graph rationalization,'' 2024. [Online]. Available: \url{https://openreview.net/forum?id=XcwHDoKvVg}
\BIBentrySTDinterwordspacing

\bibitem{yu2023mind}
J.~Yu, J.~Liang, and R.~He, ``Mind the label shift of augmentation-based graph ood generalization,'' in \emph{Proceedings of the IEEE/CVF Conference on Computer Vision and Pattern Recognition}, 2023, pp. 11\,620--11\,630.

\bibitem{yue2024cooperative}
L.~Yue, Q.~Liu, Y.~Liu, W.~Gao, F.~Yao, and W.~Li, ``Cooperative classification and rationalization for graph generalization,'' in \emph{Proceedings of the ACM on Web Conference 2024}, 2024, pp. 344--352.

\bibitem{jang2017categorical}
\BIBentryALTinterwordspacing
E.~Jang, S.~Gu, and B.~Poole, ``Categorical reparameterization with gumbel-softmax,'' in \emph{Proceedings of the 5th International Conference on Learning Representations}, 2017. [Online]. Available: \url{https://openreview.net/forum?id=rkE3y85ee}
\BIBentrySTDinterwordspacing

\bibitem{wei2023boosting}
C.~Wei, Y.~Wang, B.~Bai, K.~Ni, D.~Brady, and L.~Fang, ``Boosting graph contrastive learning via graph contrastive saliency,'' in \emph{Proceedings of the 40th International Conference on Machine Learning}, 2023, pp. 36\,839--36\,855.

\bibitem{suresh2021adversarial}
S.~Suresh, P.~Li, C.~Hao, and J.~Neville, ``Adversarial graph augmentation to improve graph contrastive learning,'' in \emph{Proceedings of the 35th Conference on Neural Information Processing Systems}, 2021, pp. 15\,920--15\,933.

\bibitem{zhao2021data}
T.~Zhao, Y.~Liu, L.~Neves, O.~Woodford, M.~Jiang, and N.~Shah, ``Data augmentation for graph neural networks,'' in \emph{Proceedings of the 35th AAAI conference on artificial intelligence}, 2021, pp. 11\,015--11\,023.

\bibitem{park2022graph}
J.~Park, H.~Shim, and E.~Yang, ``Graph transplant: Node saliency-guided graph mixup with local structure preservation,'' in \emph{Proceedings of the 36th AAAI Conference on Artificial Intelligence}, 2022, pp. 7966--7974.

\bibitem{wang2024do}
\BIBentryALTinterwordspacing
Y.~Wang, J.~Zhang, and Y.~Wang, ``Do generated data always help contrastive learning?'' in \emph{Proceedings of the 12th International Conference on Learning Representations}, 2024. [Online]. Available: \url{https://openreview.net/forum?id=S5EqslEHnz}
\BIBentrySTDinterwordspacing

\bibitem{chen2020simple}
T.~Chen, S.~Kornblith, M.~Norouzi, and G.~Hinton, ``A simple framework for contrastive learning of visual representations,'' in \emph{Proceedings of the 37th International Conference on Machine Learning}, 2020, pp. 1597--1607.

\bibitem{hu2020open}
W.~Hu, M.~Fey, M.~Zitnik, Y.~Dong, H.~Ren, B.~Liu, M.~Catasta, and J.~Leskovec, ``Open graph benchmark: Datasets for machine learning on graphs,'' in \emph{Proceedings of the 34th Conference on Neural Information Processing Systems}, 2020, pp. 22\,118--22\,133.

\bibitem{socher2013recursive}
R.~Socher, A.~Perelygin, J.~Wu, J.~Chuang, C.~D. Manning, A.~Y. Ng, and C.~Potts, ``Recursive deep models for semantic compositionality over a sentiment treebank,'' in \emph{Proceedings of the 2013 Conference on Empirical Methods in Natural Language Processing}, 2013, pp. 1631--1642.

\bibitem{rupp2012fast}
M.~Rupp, A.~Tkatchenko, K.-R. M{\"u}ller, and O.~A. Von~Lilienfeld, ``Fast and accurate modeling of molecular atomization energies with machine learning,'' \emph{Physical review letters}, vol. 108, no.~5, p. 058301, 2012.

\bibitem{xu2018how}
\BIBentryALTinterwordspacing
K.~Xu, W.~Hu, J.~Leskovec, and S.~Jegelka, ``How powerful are graph neural networks?'' in \emph{Proceedings of the 7th International Conference on Learning Representations}, 2019. [Online]. Available: \url{https://openreview.net/forum?id=ryGs6iA5Km}
\BIBentrySTDinterwordspacing

\bibitem{creager2021environment}
E.~Creager, J.-H. Jacobsen, and R.~Zemel, ``Environment inference for invariant learning,'' in \emph{Proceedings of the 38th International Conference on Machine Learning}, 2021, pp. 2189--2200.

\bibitem{luo2020parameterized}
D.~Luo, W.~Cheng, D.~Xu, W.~Yu, B.~Zong, H.~Chen, and X.~Zhang, ``Parameterized explainer for graph neural network,'' in \emph{Proceedings of the 34th Conference on Neural Information Processing Systems}, 2020, pp. 19\,620--19\,631.

\bibitem{schlichtkrull2021interpreting}
\BIBentryALTinterwordspacing
M.~S. Schlichtkrull, N.~D. Cao, and I.~Titov, ``Interpreting graph neural networks for nlp with differentiable edge masking,'' in \emph{Proceedings of the 9th International Conference on Learning Representations}, 2021. [Online]. Available: \url{https://openreview.net/forum?id=WznmQa42ZAx}
\BIBentrySTDinterwordspacing

\bibitem{menendez1997jensen}
M.~L. Men{\'e}ndez, J.~Pardo, L.~Pardo, and M.~Pardo, ``The jensen-shannon divergence,'' \emph{Journal of the Franklin Institute}, vol. 334, no.~2, pp. 307--318, 1997.

\end{thebibliography}
\vfill
\end{document}